\documentclass[10pt,twocolumn,letterpaper]{article}
\usepackage[pagenumbers]{cvpr}

\usepackage{cuted} %

\usepackage{currfile} %

\usepackage{caption} %

\makeatletter\renewcommand\LenToUnit[1]{\strip@pt\dimexpr(#1)*\p@/\unitlength\relax}\makeatother
\usepackage{pifont}
\usepackage{multirow}
\usepackage{colortbl} %
\usepackage{tabularx} %
\definecolor{oursbg}{RGB}{251,236,222} %

\usepackage{array} %
\newcommand{\cmark}{\ding{51}}
\newcommand{\xmark}{\ding{55}}

\usepackage{algorithm}
\usepackage{algorithmic}

\DeclareMathOperator*{\argmin}{arg\,min}
\DeclareMathOperator{\Exp}{Exp}

\newcommand{\gauss}{\mathcal{G}}
\newcommand{\obs}{\mathcal{O}}
\newcommand{\R}{\mathbb{R}}
\newcommand{\SE}{\mathrm{SE}(3)}

\newcommand{\render}{\mathcal{R}}
\newcommand{\da}{\Delta\mathbf{a}}

\newcommand{\bphi}{\boldsymbol{\phi}}
\newcommand{\bmu}{\boldsymbol{\mu}}
\newcommand{\bSigma}{\boldsymbol{\Sigma}}
\newcommand{\ba}{\mathbf{a}}
\newcommand{\bc}{\mathbf{c}}

\newcommand{\bI}{\mathbf{I}}
\newcommand{\bD}{\mathbf{D}}
\newcommand{\bT}{\mathbf{T}}
\newcommand{\bA}{\mathbf{A}}
\newcommand{\bR}{\mathbf{R}}

\newcommand{\bP}{\mathbf{P}}

\newcommand{\method}{BendGS}
\usepackage[section]{placeins}

\definecolor{blue}{rgb}{0,0,0}
\definecolor{revblue}{rgb}{0,0,0}
\newcommand{\rev}[1]{{\color{revblue}#1}}
\definecolor{fixred}{rgb}{0,0,0}%
\newcommand{\fix}[1]{{\color{fixred}#1}}
\definecolor{mtgred}{rgb}{0.85,0,0}

\definecolor{chkblue}{rgb}{0,0,0}
\newcommand{\chk}[1]{{\color{chkblue}#1}}
\definecolor{fbblue}{rgb}{0,0,0}
\newcommand{\fb}[1]{{\color{fbblue}#1}}
\AtBeginDocument{\crefname{figure}{Figure}{Figures}\crefname{table}{Table}{Tables}}

\renewcommand{\dbltopfraction}{0.95}
\renewcommand{\dblfloatpagefraction}{0.8}

\definecolor{cvprblue}{rgb}{0.21,0.49,0.74}
\usepackage[pagebackref,breaklinks,colorlinks,allcolors=cvprblue]{hyperref}
\usepackage{chapterbib}
\title{\method{}: Observation-Conditioned Gaussian Self-Modeling\\ for Continuum Robots}
\author{Jeong Woon Lee$^{1*}$ \qquad Doungjun Kim$^{2*}$ \qquad Namho Kim$^{1}$\\
Seungheon Chae$^{2}$ \qquad Jongwoo Kim$^{2\dagger}$ \qquad Hyoseok Hwang$^{1\dagger}$\\[0.4em]
{\normalsize $^{1}$College of Software, Kyung Hee University \qquad $^{2}$College of Engineering, Kyung Hee University}\\
{\tt\small \{everyman123, kdoungjun, shawnbest, climbheon, jongwoo.kim, hyoseok\}@khu.ac.kr}\\
{\small $^{*}$Equal contribution \qquad $^{\dagger}$Corresponding authors}}
\begin{document}
\begingroup
\raggedbottom
\maketitle
\begin{strip}
\centering
\IfFileExists{fig/fig1_teaser_v30.pdf}{%
  \includegraphics[width=\textwidth]{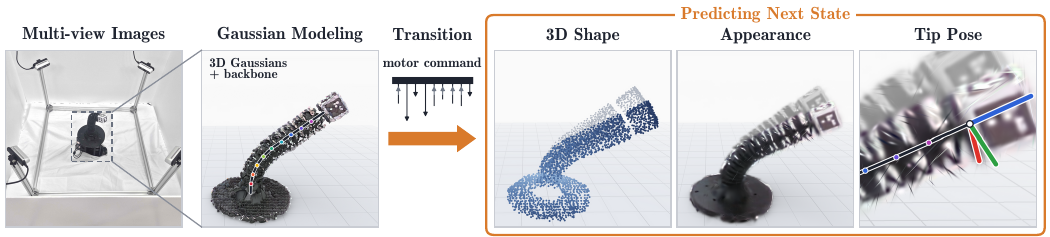}%
}{%
  \fbox{\rule{0pt}{2.2in}\rule{0.96\textwidth}{0pt}}%
}
\captionsetup{hypcap=false}
\captionof{figure}{\rev{\textbf{Visual self-modeling with \method{}.} \rev{\method{} builds a 3D Gaussian model of a continuum robot from multi-view images and predicts the next state of the robot under a relative motor command. One model provides the three outputs of the next state: 3D shape, appearance and tip pose.}}}
\captionsetup{hypcap=true}
\label{fig:teaser}%
\end{strip}

\begin{abstract}
\textcolor{blue}{Visual self-models learn from images how a robot moves under motor commands. Continuum robots need them even more, as they bend continuously without measured joint angles or fixed link geometry. Existing continuum self-models predict point clouds, image-space curves, or future images, but not the next 3D shape, appearance, and tip pose together from an observed body. 3D Gaussians offer a renderable representation of shape and appearance, yet articulated Gaussian self-models rely on joint angles to drive deformation. In this work, we propose \textbf{\method{}}, which uses a virtual backbone estimated from observation to drive command-conditioned Gaussian deformation, enabling next-state prediction for continuum robots without measured joint angles. In Gaussian modeling, a feed-forward network reconstructs the robot as 3D Gaussians from multi-view images, and the backbone, a chain of coordinate frames, is fitted to back-projected depth points to estimate the bending state. In transition, a network predicts its change under a relative motor command to deform the bound Gaussians. The deformed Gaussians render the next images and the 3D shape, and the distal frame gives the tip pose. Training losses use RGB-D images and masks, without bending or tip labels. \chk{On simulated and real robots, \method{} achieves the lowest Chamfer distance, best image quality, and lowest tip errors among the compared methods.}}
\end{abstract}

\section{Introduction}
\label{sec:intro}

\textcolor{blue}{Predicting how a robot moves in response to motor commands is central to robot control and requires reliable geometric and kinematic models~\cite{chen2022vsm,hu2026articulated,hu2025ffksm}. Conventional modeling requires human effort to build models and identify their parameters~\cite{hu2026articulated}. Self-modeling provides a data-driven alternative by learning how a robot's body responds to commands~\cite{bongard2006resilient,kwiatkowski2019taskagnostic}. Visual self-models extend this approach by learning implicit 3D models of the robot body from camera images~\cite{chen2022vsm,hu2025ffksm}. Recently, 3D Gaussian Splatting (3DGS)~\cite{kerbl2023gs} has been adopted as an explicit 3D representation for articulated robot self-modeling~\cite{hu2026articulated,liu2024drrobot}.}

\textcolor{blue}{Continuum robots need these approaches even more. Their flexible backbones bend continuously to navigate tortuous anatomy in minimally invasive surgery and confined spaces during inspection~\cite{burgnerkahrs2015survey,webster2010pcc}. Yet this flexibility makes their motion difficult to model. Unlike rigid-link robots, continuum robots lack measured joint angles and fixed link geometry that determine body shape. In widely used tendon-driven continuum robots (TDCRs)~\cite{zhang2022survey}, motors pull tendons to bend the backbone~\cite{webster2010pcc,rao2021benchmark}. The resulting shape depends on friction, gravity, and the preceding shape~\cite{cho2024hysteresis,lin2026acfm,rao2021benchmark}. Analytical approaches~\cite{webster2010pcc,rao2021benchmark,rucker2011cosserat} rely on simplifying assumptions, and their shape predictions become inaccurate when those assumptions fail.}

\textcolor{blue}{Recent studies have extended self-modeling to soft and continuum robots~\cite{cho2024hysteresis,huang2024interaction,lin2026acfm,marquesmonteiro2024visuodynamic,yu2026shapesm}. Point-cloud methods predict settled 3D body shapes from motor commands~\cite{cho2024hysteresis,lin2026acfm} without conditioning on the observed body. A curve-based method learns actuation-driven deformation from multi-view image-space curves~\cite{yu2026shapesm} and predicts tip position but not the pointing direction used in task-space kinematics~\cite{webster2010pcc}. Image-based methods predict future camera images~\cite{huang2024interaction,marquesmonteiro2024visuodynamic} without an explicit 3D body representation. These approaches do not predict the next 3D body shape, rendered appearance, and tip pose together from an observed body under a relative motor command. \rev{3D body shape supports obstacle avoidance~\cite{ataka2016pose,kasaei2025wholebody}, rendered appearance connects predicted motion to camera images~\cite{marquesmonteiro2024visuodynamic}, and tip position and pointing direction support target reaching and distal-tool alignment~\cite{loschak2017pointing}. These outputs describe different aspects of the same deformation under a motor command. A self-model should therefore predict them from one 3D representation.}}

\textcolor{blue}{3DGS provides a shared 3D representation for jointly predicting these outputs. Gaussians reconstructed from multi-view images represent the observed robot's 3D geometry and appearance in a renderable form~\cite{hu2026articulated,kerbl2023gs,liu2024drrobot}. These Gaussians deform into later shapes~\cite{luiten2024dynamic,xie2024physgaussian} and are learned through differentiable rendering~\cite{hu2026articulated,kerbl2023gs}. However, applying 3DGS to TDCRs is nontrivial. Articulated self-models obtain skeletal transforms from joint angles and deform the Gaussians with them~\cite{hu2026articulated,liu2024drrobot}. TDCRs lack measured joint angles, and tendon commands do not directly specify backbone bending~\cite{webster2010pcc,rao2021benchmark,zhang2022survey}. The key challenge is therefore to estimate the continuum robot's bending state from the observed 3D body and predict its change under tendon actuation to obtain the Gaussian deformation and tip pose.}

\textcolor{blue}{\rev{We} propose \textbf{\method{}}, a two-phase visual self-model for continuum robots\rev{, as illustrated in \cref{fig:teaser}}. In Gaussian modeling, a feed-forward network reconstructs the observed robot as 3D Gaussians from calibrated multi-view RGB images. A virtual backbone of coordinate frames~\cite{webster2010pcc,rao2021benchmark,roesthuis2016fbg} is fitted to points back-projected from rendered depth to estimate the bending state, and each Gaussian is rigidly bound to a backbone frame. In transition, a network predicts a bending-state increment from the estimated state and a relative motor command, and the updated frames deform the bound Gaussians. The next-state outputs come from the deformed Gaussians and predicted backbone. Rendering and depth back-projection produce the images and 3D shape, while the distal frame provides the tip position and pointing direction. The losses of both phases use RGB images and depth maps, without bending or tip labels. \chk{In simulation and on a real robot, \method{} achieves the lowest Chamfer distance, best image quality, and lowest tip errors among analytical and learned self-models.}}
\rev{Our contributions are:}

\begin{itemize}
  \item \textcolor{blue}{We formulate continuum-robot visual self-modeling as predicting the next 3D shape, appearance, and tip pose.}
  \item \textcolor{blue}{We propose \method{}, which fits a virtual backbone to feed-forward Gaussians to estimate the bending state and predicts its change under a relative motor command.}
  \item \textcolor{blue}{\chk{\method{} attains the lowest Chamfer distance, best image quality, and lowest tip errors among the compared methods on simulated and real robots.}}
\end{itemize}

\section{Related Work}
\label{sec:related}
\begin{figure*}[t]
  \centering
  \includegraphics[width=\linewidth]{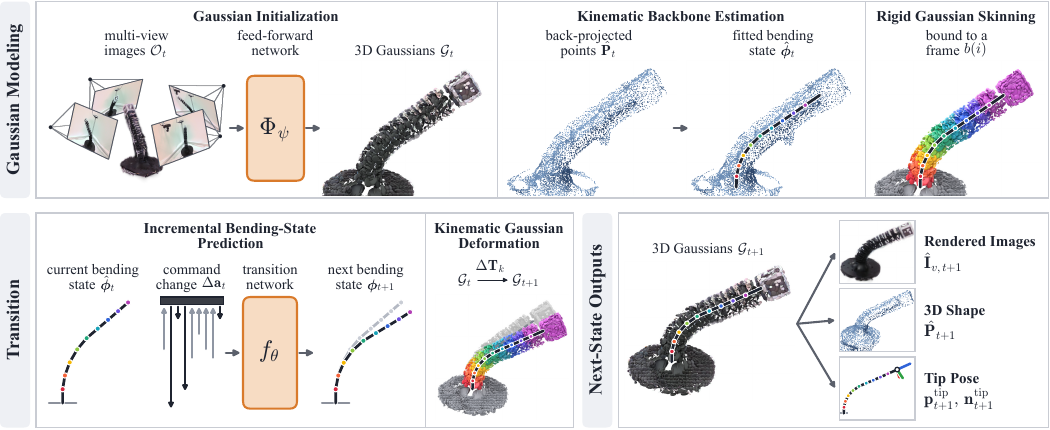}
  \caption{\textcolor{blue}{\textbf{Overview of \method{}.} A feed-forward network turns multi-view images into 3D Gaussians, and each Gaussian is bound to a frame of a fitted backbone. A transition network then maps the fitted bending state and the command change to the next bending state, and the Gaussians follow their frames. The moved Gaussians give the rendered images and the 3D shape, and the backbone gives the tip pose.}}
  \label{fig:framework}
\end{figure*}

\paragraph{Modeling continuum robots.}
\label{sec:rw_modeling}
\rev{Physics-based approaches model continuum-robot bending using constant-curvature and rod formulations~\cite{webster2010pcc,rao2021benchmark,rucker2011cosserat,till2019realtime}. Their accuracy depends on modeling assumptions and identified parameters~\cite{rao2021benchmark,dubied2022sim2real}, while physical robots exhibit nonconstant curvature~\cite{giorelli2015nn,mahl2014variable}, tendon friction~\cite{kato2015tension,wang2025friction}, hysteresis~\cite{cho2024hysteresis}, and stochastic effects~\cite{thompson2024mdn}. Sim-to-real approaches calibrate physical models or learn residual corrections~\cite{dubied2022sim2real,gao2024residual,cho2026rafl}, while simulators inherit assumptions about robot mechanics~\cite{graule2021somo,naughton2021elastica,coevoet2017sofa}. Physics-informed networks approximate rod mechanics~\cite{bensch2024pinn}. Model-less control estimates actuation-to-tip relations using online feedback~\cite{yip2014modelless,fang2019gpr}, while data-driven models learn tip or shape dynamics from measured states~\cite{bruder2021koopman,zou2026tdcr,yang2026historyfm,wu2021lstm,wang2024hysteretic}. Embedded fiber Bragg grating sensors measure the backbone shape directly but require instrumenting the robot~\cite{roesthuis2016fbg}. \method{} instead predicts the next 3D body shape and tip pose from multi-view observations and relative motor commands, without material-parameter identification or tip-pose measurements at inference.}

\paragraph{Observation-based self-models.}
\label{sec:rw_selfmodel}
\rev{Self-modeling learns a robot's body response from action-sensation data~\cite{bongard2006resilient,kwiatkowski2019taskagnostic}. For rigid-link robots, VSM and FFKSM learn implicit 3D body models~\cite{chen2022vsm,hu2025ffksm}, while articulated Gaussian self-models and differentiable robot rendering predict or render robot geometry under joint angles~\cite{hu2026articulated,liu2024drrobot}. Learned kinematic and dynamic models estimate tip responses from commands~\cite{laschi2023learning,thuruthel2017dynamic}, while visual inverse kinematics predicts the next actuation from current and target images~\cite{almanzor2023visualik}. Point-cloud methods predict settled 3D body shapes from tendon commands, using previous commands~\cite{cho2024hysteresis} or action-conditioned flow matching~\cite{lin2026acfm}. Image-driven methods predict future camera images~\cite{huang2024interaction,marquesmonteiro2024visuodynamic} or reconstruct body geometry through differentiable silhouette rendering without predicting its response to actuation~\cite{lu2023diffrender}. Among observation-conditioned approaches, Yu~\etal~\cite{yu2026shapesm} predict shape deformation from image-space curves under cable actuation and report tip position without pointing direction. NJF~\cite{li2025njf} infers per-point Jacobians from an observed image to predict 3D point motion under a command change, without explicitly providing a tip pose. \method{} combines an observed 3D body representation with relative-command-conditioned prediction of the next 3D body shape, rendered appearance, and tip position and pointing direction.}

\section{Method}
\label{sec:method}

\subsection{\rev{Gaussian Modeling}}
\label{sec:method:init}
\label{sec:method:backbone}

\paragraph{\rev{Gaussian Initialization.}}
\rev{A state is the configuration the robot settles into under a motor command; \method{} represents state $t$ by Gaussians $\gauss_t$ and predicts the shape, images and tip pose of state $t{+}1$, as \cref{fig:framework} shows. It reconstructs the robot in state $t$ as 3D Gaussians $\gauss_t=\{(\bmu_i,\alpha_i,\bSigma_i,\mathbf{sh}_i)\}_{i=1}^{N}$ (position, opacity, covariance and spherical harmonics) from the observation $\obs_t$, the images of $V$ cameras with intrinsic and extrinsic matrices $\Pi_v$. Camera $\Pi_v$ renders color and depth at pixel $u$ by compositing the Gaussians in depth order~\cite{kerbl2023gs,ye2024gsplat},}
\begin{equation}
\begin{split}
  \hat\bI_v(u)&=\textstyle\sum_{i}\bc_i(\mathbf{d}_v)\,w_i(u)\,T_i(u),\\
  \hat\bD_v(u)&=\textstyle\frac{1}{\hat\alpha_v(u)}\sum_{i}z_i\,w_i(u)\,T_i(u),
\end{split}
\label{eq:render}
\end{equation}
\rev{where the Gaussians are sorted by camera-space depth $z_i$, $\bc_i(\mathbf{d}_v)$ is the color of $\mathbf{sh}_i$ in the viewing direction $\mathbf{d}_v$, $w_i(u)$ is $\alpha_i$ weighted by the projected 2D Gaussian, $T_i(u)=\prod_{m<i}(1-w_m(u))$ is the transmittance, and $\hat\alpha_v(u)=\sum_i w_i(u)T_i(u)$ is the accumulated opacity; together, $\render(\gauss,\Pi_v)=(\hat\bI_v,\hat\bD_v,\hat\alpha_v)$. Instead of optimizing Gaussians per state as in standard 3D Gaussian splatting, \method{} predicts them with a feed-forward network $\gauss_t=\Phi_\psi(\obs_t)$; any model that maps posed multi-view images to 3D Gaussians can serve.}

\paragraph{\rev{Kinematic Backbone Estimation.}}
\rev{The backbone is a chain of frames $\bT_0,\dots,\bT_K\in\SE$, with $\bT_0$ the fixed base frame, known in the world frame of the cameras. Frame $k$ follows frame $k{-}1$ by a fixed offset $\ell_k$ along the backbone axis and a bending $\bphi_k\in\R^2$; two components suffice for an inextensible, torsion-free backbone, the usual assumption for tendon-driven continuum robots~\cite{webster2010pcc}, and the twist of a cylindrical body is not observable from its surface. With the bending state $\bphi=(\bphi_1,\dots,\bphi_K)$,}
\begin{equation}
  \bT_k(\bphi) = \bT_{k-1}(\bphi)\,\bA_k\,\Exp(\bphi_k),\quad \bA_k=\begin{bmatrix}\mathbf{I}_3&\ell_k\mathbf{e}_z\\ \mathbf{0}^{\top}&1\end{bmatrix},
  \label{eq:chain}
\end{equation}
\rev{where $\Exp(\bphi_k)\in\SE$ is the rotation by $\|\bphi_k\|$ about the axis $(\bphi_k,0)$ in the plane of the preceding frame, with zero translation. Each frame $\bT_k$ ($k\ge1$) carries a rigid cylinder $\mathcal{C}_k$ covering the body around it. The offsets $\ell_k$ and the cylinders are given by the robot model, so the bending is the only free parameter of this body model.}

\rev{The points $\hat\bP_t$ are back-projected through $\hat\bD_v$ at the pixels with accumulated opacity above $\tau_\alpha$. The chain is fitted to the body points $\hat\bP^{\mathrm{body}}_t\subset\hat\bP_t$, those outside a fixed region around the static base:}
\begin{equation}
\begin{split}
  \hat\bphi_t=\argmin_{\bphi\in\mathcal{P}}\;&\sum_{\mathbf{q}\in\hat\bP^{\mathrm{body}}_t}\rho\big(d(\mathbf{q},\bphi)\big)\\
  &+\sum_{k=2}^{K}\rho\big(\lambda_s(\bphi_k-\bphi_{k-1})\big),
\end{split}
  \label{eq:icp_update}
\end{equation}
\rev{where the first term is the data term of articulated ICP~\cite{tagliasacchi2015articulated}: $d(\mathbf{q},\bphi)=\min_{j}|\eta_j(\mathbf{q},\bphi)|$, with $\eta_j$ the signed distance (negative inside) from $\mathbf{q}$ to the surface of cylinder $\mathcal{C}_j$ at $\bT_j(\bphi)$, $j=1,\dots,K$, and $\rho$ is a robust penalty applied to each component and summed. The second term favors smooth bending along the body, and $\mathcal{P}=[-\beta,\beta]^{2K}$ bounds each component by the reachable range $\beta$.}

\paragraph{\rev{Rigid Gaussian Skinning.}}
\rev{The Gaussians follow the backbone by skinning with one-hot weights~\cite{loper2015smpl}: each Gaussian moves with one frame, chosen with the body model of the fit. In its observed pose, Gaussian $i$ is bound to the cylinder of smallest signed distance. With $\eta_j(\bmu)=\eta_j(\bmu,\hat\bphi_t)$,}
\begin{equation}
  {\color{revblue}b(i)=\textstyle\argmin_{j\in\{0,\dots,K\}}\eta_j(\bmu_i),}
  \label{eq:local}
\end{equation}
\rev{where a fixed cylinder $\mathcal{C}_0$ collects the static base. The body model also separates the robot from the scene: a Gaussian belongs to the body when it lies within $d_b$ of its cylinder and its opacity is at least $\tau_o$. The others represent the background and are removed, because they would otherwise move with the frames.}

\subsection{Transition}
\label{sec:method:transition}

\paragraph{\rev{Incremental Bending-State Prediction.}}
\rev{A transition network $f_\theta$ maps the fitted bending state $\hat\bphi_t$ and the command change $\da_t=\ba_{t+1}-\ba_t$, with $\ba_t$ the motor command under which state $t$ settles, to a bending increment of every frame:}
\begin{equation}
  \bphi_{t+1}=\hat\bphi_t+f_\theta(\hat\bphi_t,\da_t).
  \label{eq:transition}
\end{equation}
\rev{The observation \rev{provides an estimate of the current bending state}, so the network predicts only its change: with $f_\theta(\bphi,\da)=g_\theta(\bphi,\da)-g_\theta(\bphi,\mathbf{0})$ and $g_\theta$ an MLP, a zero command change leaves the bending state unchanged. The \rev{fitted bending state} and $\da_t$ are assumed to determine the next \rev{bending state}, so the absolute command is not needed and one network serves every observed state.}

\paragraph{\rev{Kinematic Gaussian Deformation.}}
\rev{Frame $k$ moves by $\Delta\bT_k=\bT_k(\bphi_{t+1})\,\bT_k(\hat\bphi_t)^{-1}$, with rotation $\Delta\bR_k$ and translation $\Delta\mathbf{t}_k$, and each body Gaussian $i$ of $\gauss_t$, with center $\bmu_{i,t}$ and covariance $\bSigma_{i,t}$, moves with its frame $k=b(i)$:}
\begin{equation}
  {\color{revblue}\begin{aligned}
  \bmu_{i,t+1}&=\Delta\bR_{k}\bmu_{i,t}+\Delta\mathbf{t}_{k},\\
  \bSigma_{i,t+1}&=\Delta\bR_{k}\bSigma_{i,t}\Delta\bR_{k}^{\top},\qquad k=b(i).
  \end{aligned}}
  \label{eq:rigid}
\end{equation}
\rev{Opacity and spherical harmonics stay fixed, as in dynamic Gaussian splatting~\cite{luiten2024dynamic} and the articulated 3DGS self-model~\cite{hu2026articulated}. The base does not move, $\Delta\bT_0=\mathbf{I}_4$, and $\gauss_{t+1}=\{(\bmu_{i,t+1},\alpha_i,\bSigma_{i,t+1},\mathbf{sh}_i)\}$ represents the next state.}

\subsection{Next-State Outputs}
\label{sec:method:rendering}
\rev{The model outputs the rendered images, the 3D shape and the tip pose of the next state. First, any camera $\Pi_v$ renders $\gauss_{t+1}$ by \cref{eq:render},}
\begin{equation}
  \rev{(\hat\bI_{v,t+1},\hat\bD_{v,t+1},\hat\alpha_{v,t+1})=\render(\gauss_{t+1},\Pi_v),}
  \label{eq:render_next}
\end{equation}
\rev{an image of the robot alone, since the background Gaussians were removed at binding. Second, the 3D shape is the point cloud $\hat\bP_{t+1}$ back-projected from $\hat\bD_{v,t+1}$ in all $V$ cameras at the pixels that the robot covers, as for $\hat\bP_t$. Third, the tip pose follows from the backbone, not from the rendering:}
\begin{equation}
  \rev{\mathbf{p}^{\mathrm{tip}}_{t+1}=\mathbf{t}_K(\bphi_{t+1}),\quad
       \mathbf{n}^{\mathrm{tip}}_{t+1}=\bR_{K-1}(\bphi_{t+1})\,\mathbf{e}_z,}
  \label{eq:tip}
\end{equation}
\rev{the origin of frame $K$ and, by \cref{eq:chain}, the pointing direction along the distal segment from frame $K{-}1$ to $K$, with $\bR_k$ and $\mathbf{t}_k$ the rotation and translation of $\bT_k$. Each prediction spans one command from an observed state.}

\begin{table*}[!t]
  \centering
  \footnotesize %
  \setlength{\abovecaptionskip}{2pt}
  \caption{\rev{Quantitative comparison of next-state prediction (test). Bold: best; underline: second. $^\dagger$Adapted to the relative command, one context camera. Grey: zero-transition reference\rev{, the measured shape and tip pose of the observed state} unchanged, not ranked.}}
  \label{tab:main}
  \setlength{\tabcolsep}{4.3pt}
  \renewcommand{\arraystretch}{1.08}
  \begin{tabular}{@{}l ccccccc c ccccccc@{}}
    \toprule
    & \multicolumn{7}{c}{\textbf{ACFM-Transition}} & & \multicolumn{7}{c}{\textbf{Real dataset}}\\
    \cmidrule(lr){2-8}\cmidrule(lr){10-16}
    & \multicolumn{2}{c}{3D body shape} & \multicolumn{3}{c}{Appearance} & \multicolumn{2}{c}{Tip pose} & & \multicolumn{2}{c}{3D body shape} & \multicolumn{3}{c}{Appearance} & \multicolumn{2}{c}{Tip pose}\\
    \cmidrule(lr){2-3}\cmidrule(lr){4-6}\cmidrule(lr){7-8}\cmidrule(lr){10-11}\cmidrule(lr){12-14}\cmidrule(lr){15-16}
    Method & CD$\downarrow$ & EMD$\downarrow$ & PSNR$\uparrow$ & SSIM$\uparrow$ & LPIPS$\downarrow$ & Pos.$\downarrow$ & Tilt$\downarrow$ &  & CD$\downarrow$ & EMD$\downarrow$ & PSNR$\uparrow$ & SSIM$\uparrow$ & LPIPS$\downarrow$ & Pos.$\downarrow$ & Tilt$\downarrow$ \\
    \midrule
    \rowcolor{gray!12}\multicolumn{16}{@{}l}{\textit{Analytical kinematics, command $\rightarrow$ shape}}\\
    PCC~\cite{webster2010pcc} & \rev{0.507} & \rev{0.470} & \rev{16.55} & \rev{0.800} & \rev{0.063} & \rev{11.06} & \underline{\rev{3.23}} &  & \rev{9.14} & \rev{2.29} & \rev{9.47} & \rev{0.749} & \rev{0.174} & \rev{45.9} & \rev{35.8} \\
    \midrule
    \rowcolor{gray!12}\multicolumn{16}{@{}l}{\textit{Learned self-models, command $\rightarrow$ shape}}\\
    VSM~\cite{chen2022vsm} & \underline{\rev{0.114}} & \rev{0.507} & \rev{17.48} & \underline{\rev{0.834}} & \underline{\rev{0.050}} & \underline{\rev{4.20}} & \rev{6.01} &  & \rev{3.81} & \rev{2.97} & \rev{10.73} & \rev{0.749} & \rev{0.164} & \rev{35.8} & \rev{23.1} \\
    Cho~\etal~\cite{cho2024hysteresis} & \rev{0.273} & \rev{0.463} & \rev{14.27} & \rev{0.729} & \rev{0.130} & \rev{6.27} & \rev{10.4} &  & \rev{2.13} & \underline{\rev{1.065}} & \rev{12.07} & \rev{0.773} & \rev{0.155} & \rev{25.7} & \rev{16.6} \\  %
    FFKSM~\cite{hu2025ffksm} & \rev{1.65} & \rev{1.18} & \rev{15.06} & \rev{0.736} & \rev{0.095} & \rev{22.7} & \rev{47.9} &  & \rev{39.1} & \rev{6.26} & \rev{7.87} & \rev{0.773} & \rev{0.183} & \rev{133.7} & \rev{148.6} \\
    Articulated 3DGS~\cite{hu2026articulated} & \rev{0.464} & \rev{1.14} & \underline{\rev{17.84}} & \rev{0.828} & \rev{0.067} & \rev{14.9} & \rev{32.4} &  & \rev{3.43} & \rev{3.63} & \rev{12.25} & \rev{0.716} & \rev{0.170} & \rev{33.6} & \rev{17.1} \\
    ACFM~\cite{lin2026acfm} & \rev{0.116} & \underline{\rev{0.341}} & \rev{17.26} & \rev{0.824} & \rev{0.062} & \rev{4.64} & \rev{6.27} &  & \underline{\rev{2.12}} & \textbf{\rev{0.982}} & \rev{12.24} & \rev{0.775} & \rev{0.149} & \rev{24.4} & \underline{\rev{16.3}} \\
    \midrule
    \rowcolor{gray!12}\multicolumn{16}{@{}l}{\textit{Learned self-models, observed state + relative command $\rightarrow$ next shape}}\\
    \color{gray}Zero transition & \color{gray}1.52 & \color{gray}0.737 & \color{gray}15.78 & \color{gray}0.762 & \color{gray}0.074 & \color{gray}22.6 & \color{gray}9.07 &  & \color{gray}0.117 & \color{gray}0.260 & \color{gray}14.18 & \color{gray}0.790 & \color{gray}0.136 & \color{gray}4.52 & \color{gray}2.20 \\
    NJF$^\dagger$~\cite{li2025njf} & \rev{0.604} & \rev{3.59} & \rev{15.63} & \rev{0.781} & \rev{0.085} & \rev{12.8} & \rev{25.8} &  & \rev{6.46} & \rev{2.02} & \underline{\rev{12.35}} & \underline{\rev{0.783}} & \underline{\rev{0.141}} & \underline{\rev{12.1}} & \rev{21.8} \\
    \rowcolor{oursbg}\textbf{\method{} (ours)} & \textbf{\rev{0.085}} & \textbf{\rev{0.323}} & \textbf{\rev{24.74}} & \textbf{\rev{0.941}} & \textbf{\rev{0.013}} & \textbf{\rev{1.47}} & \textbf{\rev{1.15}} &  & \textbf{\rev{0.159}} & \rev{1.176} & \textbf{\rev{15.83}} & \textbf{\rev{0.817}} & \textbf{\rev{0.101}} & \textbf{\rev{3.76}} & \textbf{\rev{4.16}} \\
    \bottomrule
  \end{tabular}
\end{table*}

\subsection{Optimization}
\label{sec:method:loss}

\rev{Training has two phases in sequence, both supervised by RGB images and by depth maps, which inference does not use; no bending or tip label enters a training loss.}

\paragraph{\rev{Phase 1: Gaussian modeling.}}
\rev{The pretrained feed-forward model $\Phi_\psi$ is fine-tuned on the robot with the rendering objective}
\begin{equation}
  \mathcal{L}_{\mathrm{gm}}=\mathcal{L}_{\mathrm{render}}+\lambda^{\mathrm{gm}}_{\mathrm{depth}}\mathcal{L}_{\mathrm{depth}}+\lambda_{\mathrm{bg}}\mathcal{L}_{\mathrm{bg}},
  \label{eq:init_loss}
\end{equation}
\rev{where $\mathcal{L}_{\mathrm{render}}$ is the rendering loss of the feed-forward model ($\ell_1$ and perceptual here) between the rendering composited over a background color $\mathbf{c}$, $\hat\bI_v+(1-\hat\alpha_v)\,\mathbf{c}$, and the training image, and}
\begin{equation}
\begin{split}
  \mathcal{L}_{\mathrm{depth}}&=\textstyle\frac{1}{|\mathcal{M}|}\sum_{(v,u)\in\mathcal{M}}\big|\hat\bD_v(u)-\bD_v(u)\big|,\\
  \mathcal{L}_{\mathrm{bg}}&=\textstyle\frac{1}{|\bar{\mathcal{M}}|}\sum_{(v,u)\in\bar{\mathcal{M}}}\hat\alpha_v(u),
\end{split}
  \label{eq:depth_bg}
\end{equation}
\rev{with $\bD_v$ the depth map of camera $v$, and $\mathcal{M}$ and $\bar{\mathcal{M}}$ the robot pixels with valid depth and the background pixels, given by the robot masks, which cover the body and the static base. The depth term resolves the geometry that RGB leaves ambiguous. The background term drives scene Gaussians to zero opacity, since they would otherwise enter $\hat\bP_t$ or move with the frames as spurious body. The same depth maps also supervise the per-view depth predictions of $\Phi_\psi$, along with standard scale~\cite{xie2024physgaussian} and smoothness regularizers. The parameters $\psi$ are then frozen.}

\paragraph{\rev{Phase 2: transition.}}
\rev{Training pairs are consecutive states $t$ and $t{+}1$ of the training trajectories with their $\da_t$. The Gaussians of state $t$ are moved by \cref{eq:transition,eq:rigid}, rendered by \cref{eq:render_next}, and compared in every camera with the image $\bI_{v,t+1}$ masked by the robot mask $\mathbf{M}_{v,t+1}$ and with the depth map of state $t{+}1$:}
\begin{equation}
\begin{split}
  \color{revblue}\mathcal{L}_{\mathrm{tr}}=\Big[\frac{1}{V}\sum_{v=1}^{V}&\color{revblue}\ell_1\big(\hat\bI_{v,t+1},\,\mathbf{M}_{v,t+1}\odot\bI_{v,t+1}\big)\Big]\\
  &\color{revblue}+\lambda^{\mathrm{tr}}_{\mathrm{depth}}\,\mathcal{L}_{\mathrm{depth}},
\end{split}
  \label{eq:loss_total}
\end{equation}
\rev{where $\ell_1(\cdot,\cdot)$ averages over all pixels, so Gaussians moved onto the background are penalized, and $\mathcal{L}_{\mathrm{depth}}$ is the depth term of \cref{eq:depth_bg} on $\hat\bD_{v,t+1}$ over the robot pixels of state $t{+}1$ in all cameras. Only $\theta$ is trained.}

\section{Experimental Results}
\label{sec:exp}
\begin{figure*}[!t]
  \centering
  \includegraphics[width=\textwidth]{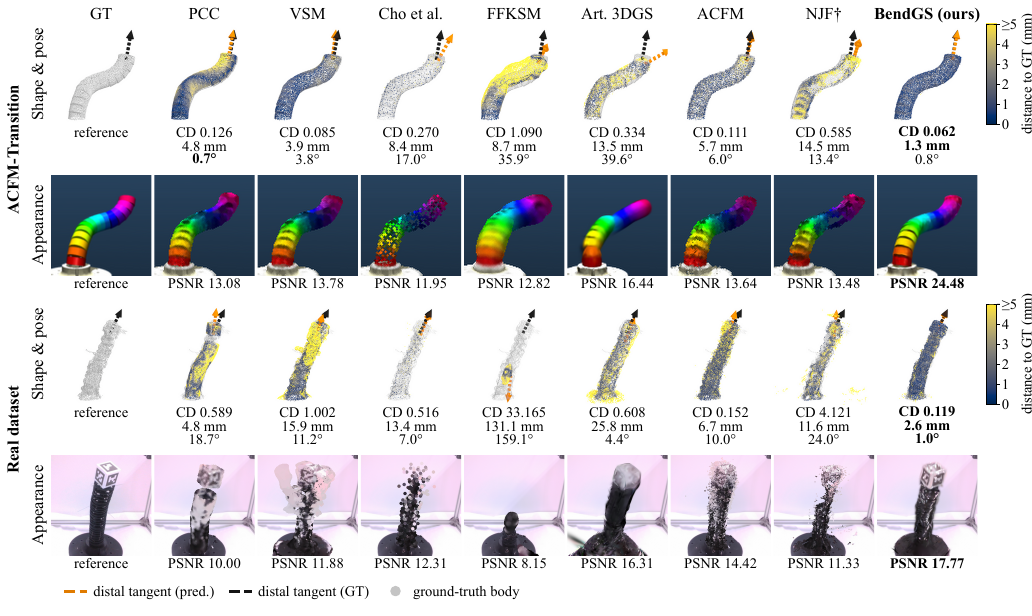}
  \caption{\rev{Qualitative comparison of next-state prediction in simulation (top) and on the real dataset (bottom). Both rows show the transition with the median tip error of \method{} within the test trajectory with the median tip error. Shape \& pose: predicted cloud coloured by the distance to the true body (grey), with predicted (orange) and true (black) distal tangents. Appearance: held-out camera. $^\dagger$As in \cref{tab:main}.}}
  \label{fig:main_qual}
  \label{fig:main_real}
\end{figure*}

\subsection{Experimental Setup}
\label{sec:exp:setup}

\begingroup\makeatletter\renewcommand\paragraph{\@startsection{paragraph}{4}{\z@}{0.6ex \@plus .2ex}{-1em}{\normalfont\normalsize\bfseries}}\makeatother
\paragraph{Datasets.}
\rev{All methods were evaluated on a simulated and a real TDCR. In simulation, ACFM-Transition keeps the robot and command range of the ACFM benchmark~\cite{lin2026acfm}, which contains no preceding state, and starts every transition from an observed state: 500 trajectories, split 400/50/50, each with ten relative commands of at most 2.5\,mm per tendon, with the eight cameras of ACFM used as inputs and sixteen held out for appearance. The real dataset comprised 50 trajectories of 100 commanded steps on a three-module TDCR with nine tendons, split 35/5/10, with four calibrated RGB-D cameras as inputs and a fifth held out. The robot masks were obtained automatically from the depth, and a marker cube on the last disc provided the reference tip pose.}

\paragraph{Metrics.}
\rev{Shape and appearance were measured on the body only. Shape was measured by the CD ($\times10^{4}$) and EMD ($\times10^{3}$) of ACFM, with the static base removed, against a ground-truth cloud back-projected from the input depth. Appearance was measured by PSNR, SSIM and LPIPS on the held-out cameras inside the ground-truth body masks of the observed and the next state, with point clouds rendered by one point-splatting rule. Tip pose was measured by the position error of the last disc and the tilt of the distal segment, against the ground-truth disc frames in simulation and the tip marker on the real robot, and was read from the backbones of \method{} and PCC and from a backbone fitted to the predicted cloud for every other method.}

\paragraph{Baselines.}
\rev{PCC~\cite{webster2010pcc} poses a constant-curvature body from the command, with parameters identified on the training split. VSM, Cho~\etal~\cite{cho2024hysteresis}, FFKSM, articulated 3DGS and ACFM map the absolute command of the next state to a shape; Cho~\etal~also received the previous command. NJF predicts motion from an image and was adapted to the relative command. Every learned baseline was retrained from released code, validated against reported results where available, and tuned on simulation validation. On the real robot, every method retained these settings with a few adaptations and selected the checkpoint on real validation.}

\paragraph{Implementation details.}
\rev{\method{} used ReSplat~\cite{xu2026resplat} as $\Phi_\psi$, initialized from the DL3DV checkpoint. The frames were placed at the spacer discs, $K{=}15$ in simulation and $K{=}9$ on the real robot. The marker cube of the real robot was bound to the last frame and treated as part of the body by every method and metric. Settings and checkpoints were selected on validation shape and tip measurements. Further details are in \cref{sec:supp:setup}.}
\par\endgroup

\subsection{Performance of Next-State Prediction}
\label{sec:exp:main}
\label{sec:exp:anystate}
\label{sec:exp:real}

\rev{In simulation, \method{} had the best mean value in every column of \cref{tab:main}. In shape, VSM and ACFM came closest, with a CD of 0.114 and 0.116 against 0.085. In appearance, our method led by 6.9\,dB in PSNR. \rev{In tip pose, the proposed method lowered the lowest baseline tip error from the 4.20\,mm of VSM to 1.47\,mm. }VSM and ACFM, the closest in CD, placed the tip within 4.64\,mm, yet their tilt of $6.01^\circ$ and $6.27^\circ$ was more than five times the $1.15^\circ$ of ours. \rev{Our method read by the fit of the baselines stayed lowest, at 1.46\,mm and $1.12^\circ$.} \rev{In the simulation row of \cref{fig:main_qual}, the clouds of FFKSM and NJF left the true body, and the tangents of Cho~\etal, FFKSM, articulated 3DGS and NJF left the true one by more than $13^\circ$.}}

\rev{On the real dataset, our method had the best mean value \rev{among the compared methods} in every column except the EMD. \rev{In the real row of \cref{fig:main_qual}, our cloud lay closest to the true body, and only our render reproduced the body and the marker cube.} ACFM and Cho~\etal~had a lower EMD, 0.982 and 1.065 against 1.176. In appearance, our method led with 15.83\,dB against at most 12.35\,dB. In tip pose, NJF had the lowest baseline tip error at 12.1\,mm, and \method{} \rev{reduced this error to 3.76\,mm} and reduced the lowest baseline tilt from the $16.3^\circ$ of ACFM to \rev{$4.16^\circ$}. \rev{Our method read by the fit of the baselines stayed lowest, at 6.09\,mm and $4.89^\circ$.} \rev{A real transition moved the tip by 4.52\,mm on average, less than the 22.6\,mm in simulation, so the zero-transition reference had a lower mean CD, EMD and tilt than every prediction, while our method had the lower tip error, 3.76\,mm against 4.52\,mm, and the better appearance in all three metrics. \Cref{fig:motion_zero}(a) therefore examined the effect of the learned transition with the test transitions grouped by the true tip motion: for motions above 5.10\,mm, our method had the lower \rev{mean} CD, 0.174 against 0.200, and the lower tip error, 4.43\,mm against 8.00\,mm, and our appearance was better in all three groups. The tilt of the reference stayed lower; our $4.16^\circ$ lay close to the $3.77^\circ$ of the Gaussians reconstructed from the true next state, so it was bounded by the reconstruction rather than by the transition. \rev{\Cref{fig:motion_zero}(b) enlarged the motion by applying \method{} $k$ times in a row without a new observation, up to ten steps, and showed the benefit of the learned transition: the tip error was 13.06\,mm against 22.43\,mm after five steps and 24.41\,mm against 42.23\,mm after ten, and \rev{the CD was lower than the reference from two steps on and the tilt from five}.} More quantitative and qualitative results are in \cref{sec:supp:additional}.}}
\begin{figure}[t]
  \centering
  \includegraphics[width=\columnwidth]{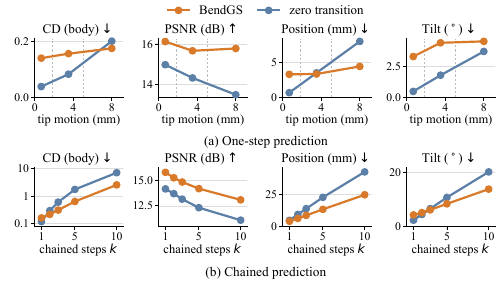}
  \caption{\rev{Comparison with the zero-transition reference on the real dataset. Dashed in (a): boundaries of the three motion groups.}}
  \label{fig:motion_zero}
\end{figure}

\subsection{Feed-Forward and Per-Scene Reconstruction}
\label{sec:exp:ff}
\leavevmode\rev{We validated in \cref{fig:ff} whether the trained feed-forward model reaches the accuracy of per-scene 3DGS optimization at a fraction of its time per state. The model took 0.22\,s per state and reached a PSNR of 27.36\,dB. In the same time, per-scene optimization reached 9.75\,dB. It needed 16\,s, 73 times longer, to match the feed-forward tip error within 0.1\,mm, and 53.1\,s, 240 times longer, for its best result. Even then its PSNR and body CD stayed behind, and only its tip pose was more accurate. Feed-forward reconstruction therefore stayed within 0.20\,mm and $0.35^\circ$ of per-scene optimization at 1/240 of its time.}

\begin{figure}[t]
  \centering
  \includegraphics[width=\columnwidth]{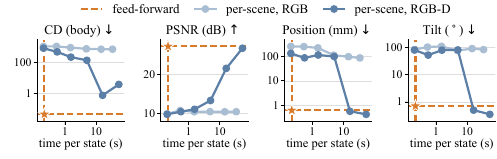}
  \caption{\rev{Accuracy against time per state for per-scene 3DGS, 30 to 7{,}000 iterations, and the feed-forward model, dashed, on two ACFM-Transition test trajectories.}}
  \label{fig:ff}
  \label{tab:ff}
\end{figure}

\subsection{Validation of Rigid Gaussian Skinning}
\label{sec:exp:backbone}
\begin{figure}[t]
  \centering
  \includegraphics[width=\columnwidth]{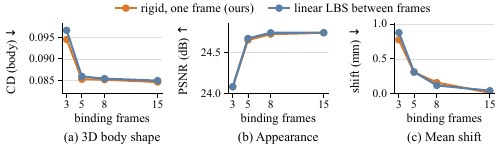}
  \caption{\rev{Comparison of rigid binding and LBS on ACFM-Transition against the number of binding frames. (c) Mean shift of the Gaussians from the positions under rigid binding with 15 frames. The tip pose does not change with the number of binding frames.}}
  \label{fig:chain_k}
\end{figure}
\leavevmode\rev{We validated in \cref{fig:chain_k} whether rigid binding to the backbone frames is expressive enough to represent the body. We compared rigid binding with linear blend skinning (LBS)~\cite{loper2015smpl,huang2024scgs} and with fewer frames. Rigid binding moves each Gaussian with one frame, and LBS moves each Gaussian with a weighted average of the neighbouring frames. With all 15 frames, the two bindings differed by 0.0003 in CD and had the same PSNR. The binding frames were then reduced without retraining. With 8 or 5 frames, the body CD of each binding rose by at most 1.2\,\%, and with 3 frames the CD rose by 12\,\% under rigid binding and 14\,\% under LBS. The error followed the number of frames and not the binding rule: LBS did not lower the CD at any frame count. Rigid binding to the 15 frames was therefore expressive enough to represent the body.}

\subsection{Analysis of the Learned Transition}
\label{sec:exp:transition_all}\label{sec:exp:transition}

\leavevmode\rev{\Cref{tab:transition} examined what the transition should output. A free motion of each frame in $\SE$~\cite{huang2024scgs} did not keep the frames on the chain and gave the highest tip error, 2.74\,mm. A bending increment linear in the command change~\cite{yip2014modelless,li2025njf} kept the frames on the chain and lowered the tip error to 2.03\,mm, but raised the CD from 0.102 to 0.111. The nonlinear increment of \method{} reduced the CD to 0.085 and the tip error to 1.47\,mm, with a tilt $0.03^\circ$ higher, and \cref{fig:motion_bins} showed that the gap to the linear increment widened as the tip motion grew from 10 to 36\,mm.
The command-to-shape baselines of \cref{tab:main} do not receive the observed state, so \cref{tab:increment} applied the increments of PCC and ACFM, the difference between the backbones fitted at the previous and at the next command, to the same observed Gaussians. The PCC increment gave a tip error of 3.23\,mm, the ACFM increment 3.59\,mm, and the learned increment 1.47\,mm. Only the increment differed between the rows, so the learned transition accounted for this reduction.}
\begin{figure}[!t]
  {  \centering
  \footnotesize
  \setlength{\tabcolsep}{1.1pt}
  \setlength{\abovecaptionskip}{2pt}
  \captionof{table}{\rev{Quantitative comparison of transition designs on the ACFM-Transition test set, with the same observation, backbone and Gaussians.}}
  \label{tab:transition}
  \resizebox{\columnwidth}{!}{%
  \begin{tabular}{@{}l cc ccc cc@{}}
    \toprule
    & \multicolumn{2}{c}{Shape} & \multicolumn{3}{c}{Appearance} & \multicolumn{2}{c}{Tip pose} \\
    \cmidrule(lr){2-3}\cmidrule(lr){4-6}\cmidrule(lr){7-8}
    Transition & CD$\downarrow$ & EMD$\downarrow$ & PSNR$\uparrow$ & SSIM$\uparrow$ & LPIPS$\downarrow$ & Pos.$\downarrow$ & Tilt$\downarrow$ \\
    \midrule
    Free $\SE$ per frame~\cite{huang2024scgs} & \underline{\rev{0.102}} & \rev{0.342} & \underline{\rev{24.57}} & \rev{0.935} & \underline{\rev{0.014}} & \rev{2.74} & \rev{2.06} \\
    Linear Jacobian~\cite{yip2014modelless,li2025njf} & \rev{0.111} & \underline{\rev{0.332}} & \rev{24.42} & \underline{\rev{0.936}} & \underline{\rev{0.014}} & \underline{\rev{2.03}} & \textbf{\rev{1.12}} \\
    \rowcolor{oursbg}Nonlinear (ours) & \textbf{\rev{0.085}} & \textbf{\rev{0.323}} & \textbf{\rev{24.74}} & \textbf{\rev{0.941}} & \textbf{\rev{0.013}} & \textbf{\rev{1.47}} & \underline{\rev{1.15}} \\
    \bottomrule
  \end{tabular}}
  }\par\vspace{6pt}
  \centering
  \begin{minipage}[t]{0.52\columnwidth}\vspace{0pt}
    \centering
    \includegraphics[width=\linewidth]{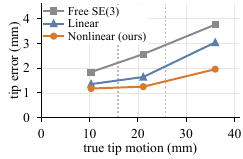}
    \setlength{\abovecaptionskip}{2pt}\par\vspace{-2pt}%
    \captionof{figure}{Tip error of the transition designs by tip motion. \rev{Dashed: group boundaries.}}
    \label{fig:motion_bins}\label{fig:increment}
  \end{minipage}\hfill
  \begin{minipage}[t]{0.47\columnwidth}\vspace{0pt}
    \centering
    \setlength{\abovecaptionskip}{2pt}
    \captionof{table}{\rev{Bending increments on ACFM-Transition.}}
    \label{tab:increment}
    \setlength{\tabcolsep}{1.0pt}
    \fontsize{7.46}{8.9}\selectfont
    \begin{tabular}{@{}l cccc@{}}
      \toprule
      Increment & CD$\downarrow$ & PSNR$\uparrow$ & Pos.$\downarrow$ & Tilt$\downarrow$ \\
      \midrule
      PCC $\Delta$ & 0.149 & 23.42 & 3.23 & 1.21 \\
      ACFM $\Delta$ & 0.102 & 23.34 & 3.59 & 5.20 \\
      \rowcolor{oursbg}\method{} & \textbf{0.085} & \textbf{24.74} & \textbf{1.47} & \textbf{1.15} \\
      \bottomrule
    \end{tabular}
  \end{minipage}
\end{figure}

\begin{table}[!t]
  \centering
  \footnotesize
  \setlength{\tabcolsep}{0.5pt}
  \setlength{\abovecaptionskip}{2pt}
  \caption{\rev{Loss ablation on ACFM-Transition: (a) Phase~1, observed state; (b) Phase~2, predicted next state.}}
  \label{tab:loss1}
  \label{tab:loss2}
  (a) Phase~1\\[1pt]
  \setlength{\tabcolsep}{1.40pt}
  \resizebox{\linewidth}{!}{%
  \begin{tabular}{@{}l ccc cc ccc cc@{}}
    \toprule
    & & & & \multicolumn{2}{c}{Shape} & \multicolumn{3}{c}{Appearance} & \multicolumn{2}{c}{Tip pose} \\
    \cmidrule(lr){5-6}\cmidrule(lr){7-9}\cmidrule(lr){10-11}
    & $\mathcal{L}_{\mathrm{render}}$ & $\mathcal{L}_{\mathrm{depth}}$ & $\mathcal{L}_{\mathrm{bg}}$ & CD$\downarrow$ & EMD$\downarrow$ & PSNR$\uparrow$ & SSIM$\uparrow$ & LPIPS$\downarrow$ & Pos.$\downarrow$ & Tilt$\downarrow$ \\
    \midrule
    (i) & \cmark & \xmark & \xmark & \rev{557} & \rev{71.9} & \rev{12.68} & \rev{0.677} & \rev{0.181} & \rev{175} & \rev{85.4} \\
    (ii) & \cmark & \xmark & \cmark & \rev{0.247} & \rev{0.391} & \textbf{\rev{27.52}} & \textbf{\rev{0.944}} & \textbf{\rev{0.009}} & \rev{4.46} & \rev{7.63} \\
    (iii) & \cmark & \cmark & \xmark & \rev{564} & \rev{72.1} & \rev{14.58} & \rev{0.727} & \rev{0.114} & \rev{154} & \rev{78.0} \\
    \rowcolor{oursbg}(iv) & \cmark & \cmark & \cmark & \textbf{\rev{0.072}} & \textbf{\rev{0.313}} & \rev{26.66} & \textbf{\rev{0.944}} & \textbf{\rev{0.009}} & \textbf{\rev{0.62}} & \textbf{\rev{0.82}} \\
    \bottomrule
  \end{tabular}
  }
  \\[4pt]
  (b) Phase~2\\[1pt]
  \setlength{\tabcolsep}{2.49pt}%
  \begin{tabular}{@{}cc cc ccc cc@{}}
    \toprule
    & & \multicolumn{2}{c}{Shape} & \multicolumn{3}{c}{Appearance} & \multicolumn{2}{c}{Tip pose} \\
    \cmidrule(lr){3-4}\cmidrule(lr){5-7}\cmidrule(lr){8-9}
    $\ell_1$ & $\mathcal{L}_{\mathrm{depth}}$ & CD$\downarrow$ & EMD$\downarrow$ & PSNR$\uparrow$ & SSIM$\uparrow$ & LPIPS$\downarrow$ & Pos.$\downarrow$ & Tilt$\downarrow$ \\
    \midrule
    \cmark & \xmark & \rev{0.088} & \textbf{\rev{0.319}} & \textbf{\rev{24.80}} & \rev{0.940} & \textbf{\rev{0.013}} & \rev{1.60} & \rev{1.20} \\  %
    \rowcolor{oursbg}\cmark & \cmark & \textbf{\rev{0.085}} & \rev{0.323} & \rev{24.74} & \textbf{\rev{0.941}} & \textbf{\rev{0.013}} & \textbf{\rev{1.47}} & \textbf{\rev{1.15}} \\
    \bottomrule
  \end{tabular}
  \par\vspace{8pt}
  \includegraphics[width=\columnwidth]{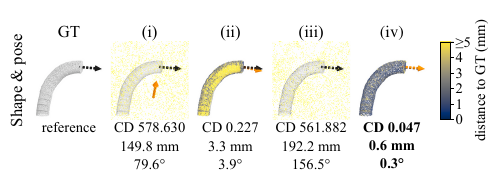}
  \setlength{\abovecaptionskip}{4pt}
  \captionof{figure}{\rev{Phase~1 reconstruction of one ACFM-Transition test state (the initialization of the transition) for the rows of \cref{tab:loss1}(a)\rev{; arrows as in \cref{fig:main_qual}}.}}
  \label{fig:loss_main}
\end{table}
\subsection{Contribution of the Loss Terms}
\label{sec:exp:losses}

\leavevmode\rev{\Cref{tab:loss1} removed the depth and the background supervision of Phase~1 and the depth supervision of Phase~2 from the objective of each phase. Without the background loss, opaque Gaussians filled the space in front of the robot and the back-projected cloud filled the workspace, as \cref{fig:loss_main}(iii) shows. Without the depth loss, the body still rendered well, with a PSNR above that of the full objective, but the reconstructed surface sank into the body, as \cref{fig:loss_main}(ii) and \cref{fig:depth_main} show: 25.9\,\% of the reconstructed points lay more than 6\,mm inside the true surface, against 1.6\,\% with the full objective. The tip error rose from 0.62 to 4.46\,mm, because the backbone is fitted to the points back-projected from the rendered depth. In Phase~2, the depth loss supervises only the transition network and lowered the tip error by 8\,\% and the CD by 4\,\%, while the EMD and the PSNR stayed nearly unchanged.}
\label{sec:exp:loss1}

\label{sec:exp:loss2}

\subsection{Sensitivity to Hyperparameters}
\label{sec:exp:sens}
\leavevmode\rev{\Cref{fig:sens} analyzed how far the result depended on the three loss coefficients, the depth and background weights of Phase~1 and the depth weight of Phase~2, each changed in turn over five values with the others fixed. The body CD varied by at most 0.006 within a sweep and stayed below the 0.114 of the best baseline in every sweep. In \cref{sec:supp:sens} of the supplementary material, the two binding thresholds changed the body CD by at most 0.005, and all seven metrics stayed better than the best baseline of each metric, except the EMD at the two smallest $\lambda_{\mathrm{bg}}$, so the lead over the baselines held across these choices.}
\begin{figure}[!t]
  \centering
  \includegraphics[width=\linewidth]{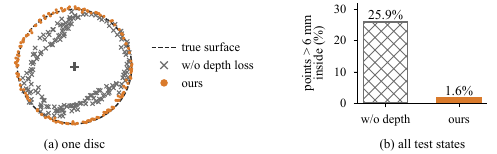}
  \caption{\rev{(a) One reconstructed disc and (b) points sunk into the body without the depth loss of Phase~1, on ACFM-Transition.}}
  \label{fig:depth_main}
\end{figure}
\begin{figure}[!t]
  \centering
  \includegraphics[width=\columnwidth]{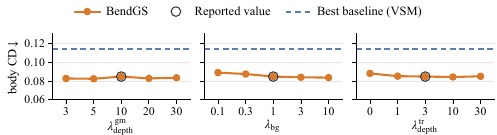}
  \caption{\rev{Test body CD on ACFM-Transition with one loss coefficient changed at a time.}}
  \label{fig:sens}
\end{figure}
\begin{figure}[!t]
  \centering
  \includegraphics[width=\linewidth]{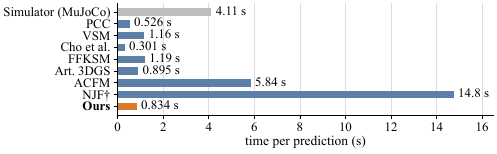}
  \caption{\rev{Time of one prediction on ACFM-Transition.}}
  \label{fig:time}
\end{figure}
\subsection{Analysis of Inference Time}
\label{sec:exp:time}

\leavevmode\rev{We measured in \cref{fig:time} the time of one prediction, from the input of a transition to the predicted body and tip pose, as the median over 50 test transitions on one B200 GPU; \rev{the baselines without frames include the backbone fit that reads the tip}. \method{} took 0.83\,s, most of it for the Gaussian reconstruction and the backbone fit; \rev{once a state was observed, the prediction for another command took 3\,ms, or 47\,ms with the body cloud}. ACFM, with 100 integration steps, was seven times slower, and NJF, the only other method that took the observed images, 18 times slower. The other command-to-shape models took 0.30 to 1.19\,s.}

{\let\FloatBarrier\relax\section{Conclusion}
\label{sec:conclusion}

\rev{We present \method{}, a visual self-model that predicts the next 3D body shape, rendered appearance, and tip pose of a continuum robot from multi-view images and a relative motor command. \method{} binds reconstructed Gaussians to a fitted virtual backbone and predicts its bending-state change. In simulation and on a real robot, \method{} achieves the lowest Chamfer distance, best image quality, and lowest tip errors among analytical and learned self-models.}

{\makeatletter\renewcommand\paragraph{\@startsection{paragraph}{4}{\z@}{0.6ex \@plus .2ex}{-1em}{\normalfont\normalsize\bfseries}}\makeatother
\paragraph{Limitations.} \rev{This work addresses visual self-modeling of continuum robots as prediction of the next 3D shape, appearance and tip pose for quasi-static free motion after one relative command from calibrated multi-view RGB images. Future work will use this prediction in closed-loop control, reduce the dependence on cameras, and extend to contact, external loads and long horizons.}\par}
}
\section*{LLM Usage Statement}
We used Claude for coding assistance, language editing, literature search, and consistency checks, and ChatGPT for manuscript restructuring, language editing, and consistency checks. Claude assisted with implementing and refactoring experiment, measurement and figure code and with running the experiments under the authors' direction, which the authors read and verified. The reported experimental measurements originate from experiment outputs. The authors are responsible for the research design, experiments, final claims, citations, numerical results, and implementation descriptions.

{
    \small
    \bibliographystyle{ieeenat_fullname}
    \bibliography{main}
}
\clearpage

\clearpage\endgroup

\clearpage
\providecommand{\mtgk}[1]{#1}
\newpage
\twocolumn[{\centering\Large\textbf{\thetitle}\\\vspace{0.5em}Supplementary Material\\\vspace{1.5em}}]
\setcounter{section}{0}
\renewcommand{\thesection}{\Alph{section}}
\renewcommand{\theHsection}{S\Alph{section}}
\setcounter{table}{0}
\renewcommand{\thetable}{S\arabic{table}}
\renewcommand{\theHtable}{S\arabic{table}}
\setcounter{figure}{0}
\renewcommand{\thefigure}{S\arabic{figure}}
\setcounter{equation}{0}
\renewcommand{\theequation}{S\arabic{equation}}
\renewcommand{\theHfigure}{S\arabic{figure}}
\makeatletter\renewcommand\@fb@secFB{}\makeatother
\raggedbottom
\setlength{\textfloatsep}{8pt plus 2pt minus 2pt}\setlength{\floatsep}{6pt plus 2pt minus 2pt}\setlength{\dbltextfloatsep}{8pt plus 2pt minus 2pt}\setlength{\dblfloatsep}{6pt plus 2pt minus 2pt}
\renewcommand{\floatpagefraction}{0.8}\renewcommand{\dblfloatpagefraction}{0.8}\renewcommand{\topfraction}{0.85}\renewcommand{\dbltopfraction}{0.85}\renewcommand{\bottomfraction}{0.5}\renewcommand{\textfraction}{0.1}\setcounter{topnumber}{3}\setcounter{bottomnumber}{2}\setcounter{totalnumber}{5}\setcounter{dbltopnumber}{3}

\section{Experimental Setup}
\label{sec:supp:setup}
\subsection{Datasets}
\label{sec:supp:datasets}
\subsubsection{Simulated Dataset: ACFM-Transition}
\label{sec:supp:data}

\begin{figure*}[t]
  \centering
  \includegraphics[width=\textwidth]{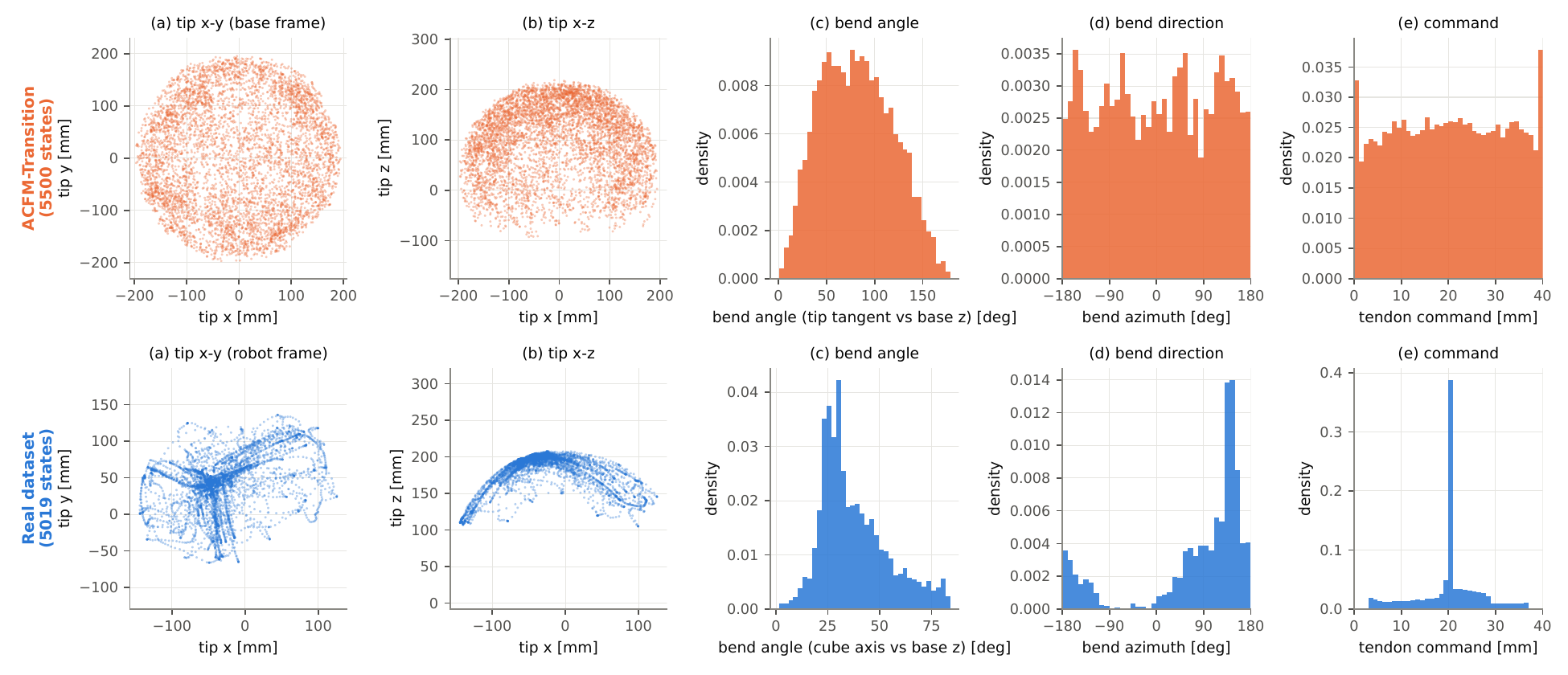}
  \caption{\rev{State and command distributions of ACFM-Transition (top) and the real-robot dataset (bottom): (a, b) tip position, (c) bend angle, (d) bend direction, (e) tendon commands.}}
  \label{fig:supp:dataset}\label{fig:supp:dataset_real}
\end{figure*}

\paragraph{Simulator and generation.}
\rev{ACFM-Transition is generated in MuJoCo 3.15.0 with the three-module model of ACFM~\cite{lin2026acfm}, unchanged, with no contact or friction. A command is the set-point of a position actuator on each tendon length, in $[0,40]$\,mm. The data are quasi-static: a state is settled when a 50{,}000-step window has $\max|\dot{\mathbf{q}}|<10^{-6}$ at every step.}

\begin{figure}[t]
  \centering
  \includegraphics[width=\linewidth]{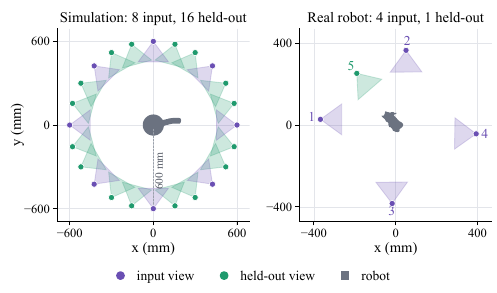}
  \caption{\mtgk{Camera layout of both datasets seen from above. The held-out views serve the appearance metrics only.}}
  \label{fig:supp:cameras}
\end{figure}

\paragraph{Rendering.}
\rev{Twenty-four cameras on a ring, $15^\circ$ apart, render $896{\times}512$ images, and the eight input cameras are $45^\circ$ apart.} \mtgk{\Cref{fig:supp:cameras} shows the layout of both datasets.}

\paragraph{Splits.}
\rev{A strongly bent disc enters the base mesh in seven training trajectories (43 states) and in none of validation or test; each of these trajectories is removed whole, and a new trajectory generated from the next unused random seed takes its place under a rule fixed before generation.}

\subsubsection{Real-Robot Dataset}
\label{sec:supp:realrobot}
\label{sec:supp:real}
\begin{figure}[t]
  \centering
  \includegraphics[width=0.62\linewidth]{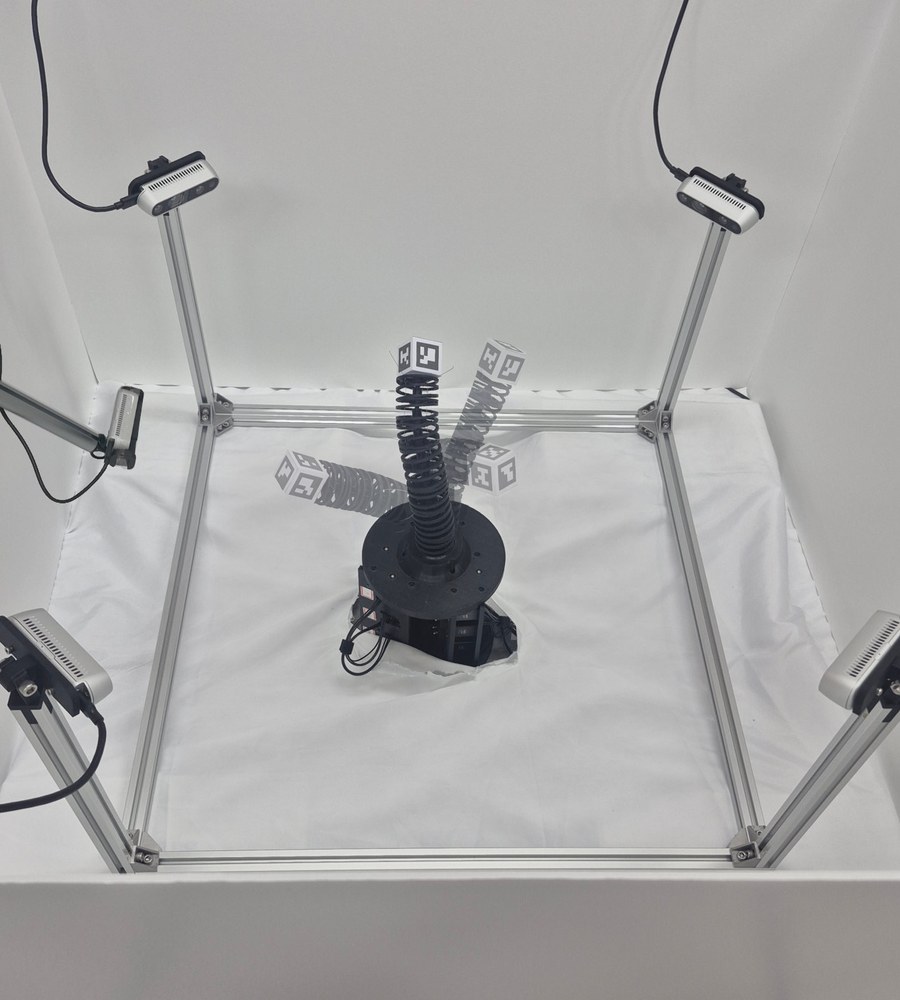}
  \caption{\fb{Real robot: four poses overlaid at the base. The background was edited for visibility.}}
  \label{fig:supp:hardware}
\end{figure}

\paragraph{Robot \fb{and hardware}.}
\rev{The real robot was a three-module TDCR with nine rings, three per module\fb{; its modules measured 65, 65 and 60\,mm}\fb{, and \cref{fig:supp:hardware} shows the robot}. Each module was bent by three tendons, driven by position-controlled Dynamixel motors\fb{, and the three tendon increments of one module summed to zero in every step~\cite{grassmann2024clarke}}. A command is the displacement of each tendon from the home position, limited to $\pm16.6$\,mm from home and to 1.38\,mm per step; \cref{fig:supp:dataset} plots it as 20\,mm minus the pull. One common factor on the CAD ring spacings (1.017) and the mount offset (23\,mm) and yaw of the marker cube were calibrated without labels on the training states: the factor on the ground-truth depth clouds, and the mount on the reconstructions of \method{}.}

\paragraph{Camera setup.}
\rev{Five RealSense RGB-D cameras surrounded the robot, as \cref{fig:supp:cameras} shows. Cameras 1 to 4 are the input cameras and camera 5 is the held-out camera. Camera 5 was registered per state from background features with the robot masked out, and its fixed offset and the depth bias of each camera were fitted on the 35 training trajectories. The robot frame had its $z$ axis along the normal of the top of the base disc and its origin at the centre of that disc.}

\paragraph{Collection.}
\rev{Each trajectory held 101 settled states and moved one to six of the six bending degrees of freedom of the robot. Every trajectory started from and returned to the home command, and a step changed the nine tendon lengths by a median of 1.20\,mm (L2 norm). A state was captured after a wait of 1.5\,s, once every motor had stayed within three encoder ticks for 0.2\,s. Every method received the commanded tendon lengths and never the measured positions.}

\paragraph{Splits and transitions.}
\rev{The trajectories were split before any result was computed, stratified by the number of moved degrees of freedom, with seed 0. The 31 states with motor motion during the capture or a marker reprojection error above 5\,px were excluded before training with every transition that contains them, which leaves 494 validation and 996 test transitions.}

\paragraph{Ground truth.}
\rev{The tip pose was the pose of a 30\,mm marker cube on the last ring, estimated from its marker corners seen by cameras 1 to 4, and the tip point was the centre of the cube face attached to the ring. The marker corners reprojected with a median RMS error of 1.92\,px, and on the stationary robot the tip measured at the end of one trajectory and at the start of the next agreed to a median of 0.11\,mm. The ground-truth shape was the body region of \cref{sec:supp:bodymask} in the cloud back-projected from the corrected depth of cameras 1 to 4, which covers the visible surface only. The tilt was the angle between the predicted direction and the pointing axis of the labelled cube. Tip labels entered no training loss of any method.}

\subsection{Evaluation Metrics and Protocol}
\label{sec:supp:toc_eval}
\label{sec:supp:metrics}
\subsubsection{\fix{Body Region and Robot Masks}}
\label{sec:supp:bodymask}
\begin{figure}[!tb]
  \centering
  \includegraphics[width=\linewidth]{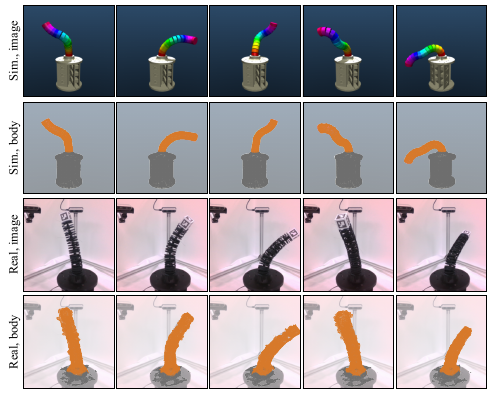}
  \caption{\fix{Scored body region on five test states of ACFM-Transition (top two rows) and the real robot (bottom two rows). Orange: scored body; grey: removed base.}}
  \label{fig:supp:body10}
\end{figure}

\fix{Every method is scored against one body cloud and one set of robot masks per dataset, shown in \cref{fig:supp:body10}. \method{} is supervised by the depth maps and masks of the same points.}

\rev{In simulation, the body cloud is back-projected from the depth of the rendered robot pixels in the eight input cameras. On the real robot, every depth pixel of an input camera was back-projected into the robot frame of \cref{sec:supp:real}; the points inside the cylinder $3<z<260$\,mm and $r<230$\,mm, with $r$ the distance from the base axis and $z$ the height above the base disc, form the body cloud, and their pixels form the body mask of each camera.}

\rev{The shape metrics remove the static base from the ground truth and from every prediction: the region $\mathcal{R}_0$ of $r<85$\,mm with $z<180$\,mm, or $z\le100$\,mm, in simulation, and the region $r<85$\,mm with $z<16$\,mm on the real robot. The scoring also keeps only the largest connected component of the real ground-truth cloud, with components linked at 10\,mm.} \fix{A prediction is cut by the same regions and passes no component filter, so the points that a method places off the body stay in its cloud and raise its error. These regions were fixed before any method was scored.}

\fix{The body mask of the held-out camera 5, the region of the real appearance metrics, came from the body cloud projected into camera 5.}

\subsubsection{Metrics}
\rev{Shape was measured by the Chamfer distance (CD) and the earth mover's distance (EMD), as in ACFM. Both clouds are voxelized at 2\,mm and 10{,}000 points are sampled from each with fixed seeds. CD is the squared two-sided Chamfer distance and EMD the approximate matching cost of the ACFM code. CD-full keeps the static base and serves only the binding and checkpoint selections of \cref{tab:supp:hparams}, and CD-body removes it. In simulation, tilt is the angle between the predicted and the true chord from the centre of disc $K{-}1$ to that of disc $K$, for every method.}

\rev{Appearance was measured by the peak signal-to-noise ratio (PSNR), the structural similarity (SSIM) and the learned perceptual image patch similarity (LPIPS). The rendered next state is composited over a background image, the per-pixel median of the training states over the pixels the robot does not cover. Each image is cropped to the bounding box of the robot masks of the training states and resampled to $128{\times}128$. PSNR is computed over the ground-truth body masks of the observed and the next state, and SSIM and LPIPS on their bounding box with every pixel outside the region set to black in both images. Point clouds are rendered as discs with a radius chosen on validation for each method and colours from the nearest point of the observed state. In simulation the ground-truth cloud rendered by this rule reaches a PSNR of 17.67\,dB, an SSIM of 0.837 and an LPIPS of 0.048, a reference for the point-cloud rows. Articulated 3DGS~\cite{hu2026articulated} and \method{} render their own Gaussians.}

\subsubsection{Tip Pose Readout}
\label{sec:supp:icp}
\rev{The learned baselines output point clouds without frames, so their tip pose was read by the backbone fit of \method{}. In simulation, before any baseline was scored, the fit was applied to the ground-truth cloud reduced to the point budget of each baseline and recovered the true tip within 0.12\,mm and $0.19^\circ$ for 4{,}224 points or more, and within 0.27\,mm and $0.47^\circ$ at the 512 points of Cho~\etal~\cite{cho2024hysteresis}, as listed in \cref{tab:supp:floor}. The table also gives the CD and EMD of each reduced cloud, the error that the ground truth itself received at that number of points. These values are references and not lower bounds. Read with the same fit, the predicted cloud of \method{} gave a tip error of 1.46\,mm and a tilt of $1.12^\circ$, against 1.47\,mm and $1.15^\circ$ from its own frames.}

\begin{figure}[!tb]
  \centering
  \includegraphics[width=\linewidth]{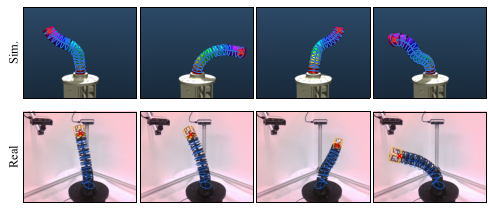}
  \caption{\rev{Backbone fit to the ground-truth body cloud on four test states of ACFM-Transition (top) and four validation states of the real robot (bottom). Blue: fitted rings and chain; orange: read tip, with the cube on the real robot; red cross: ground-truth tip.}}
  \label{fig:real_fit}
\end{figure}

\paragraph{Real robot.}
\label{sec:supp:real_readout}
\rev{On the real robot one fit read the tip pose of every predicted cloud without frames; the tip of \method{} in \cref{tab:main} is its direct output of \cref{eq:supp:tip_real}, and the tip of PCC comes from its model frames. The data term of the fit is the distance to a tube of radius $r$ around the polyline through the ring centres or to the box of the marker cube, whichever is nearer,
$d(\mathbf{q},\bphi)=\min\big(\big|\mathrm{dist}(\mathbf{q},\mathrm{polyline})-r\big|,\,|\eta_{\mathrm{cube}}(\mathbf{q})|\big)$,
with a coverage term of weight 3.0, a prior of weight 0.3 on every bending and a smoothness of weight 1.5 between the bendings inside one module. The radius $r$ is a variable of this fit, bounded to $[12,24]$\,mm, every residual passes a soft-$\ell_1$ penalty of scale 2\,mm, and the fit keeps the lowest cost of four yaw starts. The fit uses the calibrated ring spacing and cube mount offset of \cref{sec:supp:real}. The last link of the readout chain and the cube axis coincide, and the tip point is the origin of the last ring advanced by the 23\,mm mount offset along this axis. The backbone fit of \method{} on the real robot, which gives the observed bending state, minimizes \cref{eq:icp_update} as in simulation, with nine ring cylinders and the box of the marker cube.}

\rev{On the ground-truth body clouds of the 996 test transitions, the readout placed the tip 7.89\,mm and $6.03^\circ$ from the marker label on average. This value is a reference and not a lower bound, and it is subtracted from no method. The weights were defaults set before the real dataset was collected; with each changed in turn on the validation split, the ranking of all methods was unchanged and no tip error changed by more than 0.25\,mm. A mount offset refitted on the ground-truth depth clouds of the training states, 22\,mm, left the ranking unchanged and changed the direct output of \method{} in \cref{tab:main} from 3.76 to 4.10\,mm. \Cref{fig:real_fit} shows the fit on both datasets.}

\begin{table}[t]
  \centering
  \scriptsize %
  \setlength{\tabcolsep}{3pt}
  \caption{\rev{Reference for each point budget: the ground-truth cloud reduced to the number of points of each method and scored against the full ground truth on the ACFM-Transition test set.}}
  \label{tab:supp:floor}
  \begin{tabular}{@{}l r cccc@{}}
    \toprule
    Budget of & Points & CD-body & EMD-body & \shortstack{Tip\\(mm)} & \shortstack{Tilt\\($^\circ$)} \\
    \midrule
    Cho \etal & 512 & 0.204 & 0.478 & 0.27 & \rev{0.47} \\
    NJF & 4{,}224 & 0.024 & 0.207 & 0.12 & \rev{0.19} \\
    ACFM & 20{,}000 & 0.024 & 0.218 & 0.12 & \rev{0.19} \\
    \method{} & \rev{40{,}909} & 0.013 & 0.173 & 0.12 & \rev{0.19} \\
    PCC & 51{,}163 & 0.007 & 0.133 & 0.12 & \rev{0.19} \\
    \bottomrule
  \end{tabular}
\end{table}

\subsubsection{Protocol}
\label{sec:supp:protocol}
\rev{All methods were trained and evaluated under one protocol; \cref{sec:supp:baselines} gives the recipe of each baseline and every difference between a paper and its released code.}

\paragraph{Data and inputs.}
\rev{The validation split served only the choices listed below, and ACFM-Static, the published benchmark of ACFM, served only the port check of \cref{sec:supp:baselines}. In simulation, Cho~\etal~and NJF~\cite{li2025njf} attach the static base as a template, and the other methods output the base with the body. NJF received the image of the first context camera with the relative command.}

\paragraph{Checkpoints.}
\rev{Every baseline followed the checkpoint rule of its paper or released code listed in \cref{tab:supp:recipes}, with three exceptions: ACFM selects its EMA weights on the full validation split instead of the first batch of the released code, VSM keeps its last epoch as \cref{tab:supp:recipes} states, and NJF states no rule, so the lowest value of its own validation objective is used. \fix{Phase~1 uses the refinement checkpoint with the lowest validation rendering loss;} the Phase~2 checkpoint within a run is chosen, as for ACFM, by the Chamfer distance to the ground-truth clouds of the validation split, as \cref{tab:supp:hparams} specifies.}

\paragraph{Choices made on validation.}
\rev{No label enters the training loss of either phase of \method{} in \cref{eq:init_loss,eq:loss_total}: the losses use the RGB images, depth maps and robot masks of the input cameras. The tip labels of the validation split serve the selections stated below, and those of the test split serve the evaluation. The held-out cameras serve the evaluation of appearance, and inference takes the calibrated images of the observed state and the relative command. The initialization length and the depth weight of Phase~1 were selected among eight preset candidates by the validation tip error of the chain fit, and the transition settings among 18 runs. The grid was not extended after its results were seen. The other values chosen on validation, each with a rule fixed before the first result, are the binding thresholds $\tau_o$ and $d_b$, the width and depth weight of $f_\theta$, the backbone fitter and, for baselines, any value the data forced, given in \cref{sec:supp:baselines}. The threshold $\tau_\alpha$ is the Otsu threshold of the training renders and uses no label. All other weights, including $\lambda_{\mathrm{bg}}$, were fixed in advance. On the real robot, the Phase~2 checkpoint of \method{} is the latest one within 1\,\% of the lowest validation body CD, and its observation passed a validation check of the fitted tip fixed in advance, 2.27\,mm against a threshold of 5\,mm, before Phase~2.}

\paragraph{Seeds, test evaluation and statistics.}
\rev{Every row uses seed 0, and \cref{sec:supp:stat} repeats the transition network of \method{} with two further seeds. Each selected setting was evaluated on the test split once, with the one exception of Cho~\etal\ stated in \cref{sec:supp:baselines}. The other runs whose values this supplement reports serve as reference only and took no part in any selection. Confidence intervals are 95\,\% paired bootstrap intervals over the test trajectories with 10{,}000 resamples. All CDs and EMDs come from one script applied to the saved prediction clouds.}

\subsection{Baselines}
\label{sec:supp:baselines}

\paragraph{Principle.}
\rev{Apart from ACFM itself, no baseline code or training setting has been released for this robot: ACFM reports three of the baselines on the same TDCR without their training length, checkpoint or readout. Every learned baseline was therefore re-implemented from the official code of its own paper with the recipe of \cref{tab:supp:recipes} and the adaptations and validation choices stated below, under the common protocol of \cref{sec:supp:protocol}.}

\paragraph{Choices in favour of the baselines.}
\mtgk{The port made the following choices in favour of the baselines. Cho~\etal\ were trained on a body-only target of 512 points in place of the whole surface. FFKSM received eight calibrated cameras in place of one and the better of the positional-encoding values of its paper and code. Articulated 3DGS ran the better of its paper and code values. ACFM selected its checkpoint on the full validation split in place of the first validation batch of its code. In simulation, NJF received the ground-truth flow in place of RAFT flow, the ground-truth background depth, one fixed context view and half of its rays on the robot from the ground-truth mask.}

\begin{table}[t]
  \centering
  \scriptsize %
  \setlength{\tabcolsep}{3.2pt}
  \caption{\rev{Port check on ACFM-Static under ACFM's evaluation protocol. Reported: Table~I of ACFM. $^\ast$Solid occupancy readout, with the points below the floor removed.}}
  \label{tab:supp:portcheck}
  \begin{tabular}{@{}l cc cc@{}}
    \toprule
    & \multicolumn{2}{c}{CD$\times10^4$} & \multicolumn{2}{c}{EMD$\times10^3$} \\
    \cmidrule(lr){2-3}\cmidrule(lr){4-5}
    Method & Reported & Ours & Reported & Ours \\
    \midrule
    VSM~\cite{chen2022vsm} & 32.494 & 9.009 & 7.386 & 4.118 \\
    FFKSM~\cite{hu2025ffksm} & 0.974 & 1.172$^\ast$ & 2.035 & 2.915$^\ast$ \\
    Articulated 3DGS~\cite{hu2026articulated} & 1.205 & 0.684 & 4.837 & 2.763 \\
    ACFM~\cite{lin2026acfm} & 0.164 & 0.151 & 0.578 & 0.597 \\
    \bottomrule
  \end{tabular}
\end{table}

\begin{table*}[t]
  \centering
  \scriptsize
  \setlength{\tabcolsep}{2.4pt}
  \caption{\rev{Recipes of the baselines, run from their released code. Last column: values chosen on the validation split of ACFM-Transition and kept on the real robot.}}
  \label{tab:supp:recipes}
  \newcolumntype{R}[1]{>{\raggedright\arraybackslash\leavevmode}p{#1}}
    \begin{tabular}{@{}R{1.8cm} R{2.9cm} R{3.8cm} R{2.9cm} R{2.9cm} R{2.0cm}@{}}
    \toprule
    Method & Input & Training & Checkpoint & Readout & Chosen on val. \\
    \midrule
    VSM~\cite{chen2022vsm} & absolute command, scaled to $[-1,1]$ & surface points and normals, Adam $5{\times}10^{-5}$, 564k steps (the paper's 2{,}000 epochs at its batch size and data amount) & last epoch (preferred on the validation split of ACFM-Transition over epoch 2{,}000) & marching cubes, largest component & \rev{$\omega_0{=}3$} \\
    Cho~\etal~\cite{cho2024hysteresis} & previous and target absolute commands & 512-point body cloud, batch 32, Adam 0.01, $\times0.1$ every 100 epochs & lowest validation loss, early stop after 50 epochs & output cloud with base template & input scaling, \rev{lr $3{\times}10^{-3}$, $\lambda{=}0.3$} \\
    FFKSM~\cite{hu2025ffksm} & absolute command, mapped to the joint-angle range & 100$\times$100 silhouettes of 8 cameras, Adam $5{\times}10^{-4}$, up to 400k iterations & lowest validation MSE, released early stop & density threshold on a 2\,mm grid, marching cubes & \rev{5 frequencies, width 256, threshold 0.4} \\
    Articulated 3DGS~\cite{hu2026articulated} & absolute command & RGB and masks of 8 cameras, 600k iterations, AdamW $10^{-4}$ cosine & final iteration & depth fusion of the 8 rendered views & \rev{16 bones, our staging} \\
    ACFM~\cite{lin2026acfm} & normalized absolute command & 20{,}000-point clouds, authors' checkpoint settings & EMA weights by validation CD (full split) & Heun sampling, 100 steps & \rev{Hybrid, lr $1.2{\times}10^{-3}$} \\
    NJF~\cite{li2025njf} & RGB image of the first context camera (one image, as in the paper), relative command & stage A 60k steps; stage B from the 50k stage-A checkpoint, a 30k-step probe, continued to 100k only if the validation flow error fell below 0.9 of the zero-motion error, stopped at a 2\,\% plateau over 30k steps after 60k & lowest validation objective & expected surface points displaced by the predicted motion & \rev{context camera fixed (first context view)} \\
    \bottomrule
  \end{tabular}
\end{table*}

\begin{table*}[t]
  \centering
  \scriptsize %
  \setlength{\tabcolsep}{3pt}
  \caption{\rev{Validation tuning of the baselines on ACFM-Transition. Cand.: number of candidates. CD: validation body CD at the published and at the selected setting.}}
  \label{tab:supp:tuning_grid}
  \begin{tabular}{@{}>{\raggedright\arraybackslash}p{1.6cm} >{\raggedright\arraybackslash}p{4.7cm} >{\raggedright\arraybackslash}p{4.75cm} c >{\raggedright\arraybackslash}p{1.9cm} c@{}}
    \toprule
    Method & Tuned quantity and candidates & Reason for the quantity & Cand. & Selected & CD published / tuned \\
    \midrule
    Cho~\etal & loss weight $\lambda\in\{0.3,1,3\}$ of $C+\lambda E$ times the initial learning rate $\{3{\times}10^{-3},10^{-2}\}$; batch 32 and the decay schedule of the paper fixed & $\lambda$ is the only value the paper states was set experimentally (Sec.~IV, V-C) & six & lr $3{\times}10^{-3}$, $\lambda{=}0.3$ & 0.395 / 0.316 \\
    ACFM & backbone, each with the batch and learning rate of its released configuration scaled by $\{0.5,1,2\}$: Hybrid, batch 16 as in the released checkpoint, $\{3,6,12\}{\times}10^{-4}$; MLP, batch 8 as in the README, $\{1.5,3,6\}{\times}10^{-4}$ & the backbone is the only ablation of the paper (Tables I--III) & six & Hybrid, lr $1.2{\times}10^{-3}$ & 0.146 / 0.127 \\
    VSM & sine frequency $\omega_0$ of the command branch $\{30,10,3\}$; weight decay of $10^{-4}$ at $\omega_0{=}30$ & memorization of the training commands: at the published $\omega_0{=}30$, the port reached a body CD of 0.261 on training transitions and 52.2 on validation & four & $\omega_0{=}3$ & 52.2 / 0.163 \\
    FFKSM & positional-encoding frequencies $\{5,7,10\}$; network width 256 at five frequencies; density threshold over $\{0.2,\dots,0.7\}$, read on validation without retraining & five frequencies in the paper, ten in the code; a first threshold grid from 0.4 to 0.9 ended at its lower bound & 24 & 5 frequencies, width 256, threshold 0.4 & 1.86 / 1.60 \\
    Articulated 3DGS & number of bones $\{10,16\}$ times the staging of the training states $\{$ours, the paper's ratio of 500/500 images$\}$ & the paper names the number of bones an important hyperparameter; the port and the paper differ in the staging & four & 16 bones, staging of the port & 0.612 / 0.506 \\
    NJF & four runs of the port made before the tuning, evaluated on validation, and a run with a fixed context camera, evaluated from that camera & single-image setting of the paper; the official code expresses every point in the frame of the context camera, and the eight cameras form a symmetric ring; the fixed context camera needs no change to the model code & five & context camera fixed (first context view) & 1.27 / 0.639 \\
    \bottomrule
  \end{tabular}
\end{table*}

\paragraph{Verification of the port.}
\rev{Every method with a published result on the ACFM data was checked under the protocol of ACFM in \cref{tab:supp:portcheck}. ACFM-Static was regenerated with the authors' collection pipeline and arguments, and ACFM reproduced its reported values. The ports of VSM and Articulated 3DGS reached lower errors than the reported values, and the port of FFKSM reached 1.20 times the reported CD and 1.43 times the reported EMD; the settings behind the reported numbers are not stated. \method{} is not evaluated on ACFM-Static, where a sample has no observed state and no commanded change.}

\paragraph{Validation tuning on ACFM-Transition.}
\rev{Every learned baseline was tuned on the validation split of ACFM-Transition over the candidates of \cref{tab:supp:tuning_grid}, which contain the values of its paper or code. The candidates were fixed in advance and no training grid was extended after its results were seen, with two exceptions that the table states: the second threshold grid of FFKSM and the earlier runs of NJF. Each candidate kept the checkpoint rule of \cref{tab:supp:recipes}, and the candidates were compared by the body CD on the 500 validation transitions. The tuning lowered the validation body CD of every baseline, and that of VSM fell from 52.2 to 0.163.}

\paragraph{\fix{Protocol on the real robot.}}
\label{sec:supp:real_protocol}
\fix{The real-robot comparison followed one rule for every method, \method{} included.} \rev{Each method was trained on the real training split with the settings selected on the validation split of ACFM-Transition, apart from the three adaptations stated below. The real validation split selected the checkpoint of each method, by the rule of \cref{tab:supp:recipes} for the baselines and by the validation Chamfer distance for \method{}, as \cref{sec:supp:protocol} states, and the splat radius of each point-cloud output by the rule used in simulation. PCC was fitted to the training split alone. The training length of ACFM and VSM was set to the number of optimization steps of their simulation runs, since the real training split holds fewer states. The learning rate of ACFM was selected on the real validation split between the published value and the simulation-tuned value, and the simulation-tuned value was selected. For \method{}, the target images of the photometric loss of Phase~1 carry a random background colour outside the robot mask, with the input images unchanged. This variant was selected on a separate recording of the same robot and used unchanged on the dataset of this paper. The changes that the real data required of individual methods are listed in the next paragraph.}

\paragraph{Changes required by the data.}
\rev{The robot has nine tendon commands and no joint angles, so every command-conditioned network takes a nine-dimensional input, normalized as its code normalizes joint angles. The eight calibrated cameras of the simulated rig and the four input cameras of the real robot replace the camera setups of the papers, at the resolution of each method. Where a paper and its released code differ, the better of the two on validation is used; Cho~\etal\ are the exception, with the schedule of the paper and the normalization layer of the code.}
\begin{itemize}
  \item \rev{\textbf{PCC.} On the real robot, PCC \fix{gave its tip pose from its model frames, as in simulation, and} was fitted to the body clouds of the training states instead of the disc centres, with the tendon azimuths, the yaw of the marker cube and the coupling of the tendons that pass through lower modules added to its parameters.}
  \item \rev{\textbf{VSM.} Training points and normals are back-projected from the eight masked depth maps rather than estimated on the released clouds.}
  \item \rev{\textbf{Cho~\etal.} The EMD extension is a PyTorch port of the same kernels. On the real robot, the target clouds were cut at the 16\,mm base height of the evaluation. A first run with a 6\,mm cut had been tested before this mismatch was found, and \cref{tab:main} reports the retrained run.}
  \item \rev{\textbf{Articulated 3DGS.} The rasterizer is gsplat, and the static stage uses the training state with the smallest command, since the data has no zero-command state. On the real robot, the released schedule was run to completion.}
  \item \rev{\textbf{ACFM.} Clouds are back-projected from the stored depth maps of the robot mask. On the real robot, ACFM was trained on the body clouds with the base removed, the definition of the evaluation.}
  \item \rev{\textbf{NJF.} In simulation, NJF received the ground-truth flow and background depth, with mask-based ray sampling since the body covers 2\,\% of the image. On the real robot, the flow was estimated by RAFT, and NJF received zero flow on background pixels.}
\end{itemize}

\paragraph{Deviations from the published protocols.}
\rev{Differences from the protocol of ACFM (Sec.~IV-B and IV-C) and from the baseline papers:}
\begin{itemize}
  \item \rev{\textbf{Ground-truth cloud.} ACFM's released collection script back-projects the full robot, tendons included, into 20{,}000 points. On ACFM-Transition the foreground is the discs and the base without the tendons, at 896$\times$512; the check of \cref{tab:supp:portcheck} uses the released definition.}
  \item \rev{\textbf{ACFM.} The paper fixes 500 epochs and the validation CD for model selection but not the batch size, optimizer or seed. The row of \cref{tab:main} used the released checkpoint arguments (batch 16, 4{,}000 warm-up steps, cosine schedule) with the tuned learning rate of \cref{tab:supp:tuning_grid}, and trained for 455 epochs, 125{,}125 optimization steps against the 125{,}000 of the 500 epochs of its paper on its 4{,}000 training states.}
  \item \rev{\textbf{Articulated 3DGS.} Five values differ between the paper and the released code; the released values are run, with the number of bones selected on validation, and a second run with the paper values was worse on every body metric. The static reconstruction uses the eight rig views in place of the 100 views of the paper.}
  \item \rev{\textbf{FFKSM.} The main table reports the paper recipe at the tuned width and threshold of \cref{tab:supp:tuning_grid}. \fix{The empty predictions of FFKSM on the real robot are reported in \cref{tab:supp:ffksm_subset}.}}
\end{itemize}

\paragraph{Compute.}
\rev{Training times on NVIDIA B200 GPUs, several runs sharing one GPU: VSM about 22\,h, Cho~\etal~0.5 to 1\,h, FFKSM about 22\,h, Articulated 3DGS about 26\,h, ACFM 9 to 29\,h, and NJF 16 to 20\,h for both stages.}

\paragraph{PCC.}
\rev{PCC~\cite{webster2010pcc,jones2006kinematics} used three torsion-free constant-curvature sections of five joints each, and the bend of section $s$ follows the geometric tendon relation for three tendons at $0/120/240^\circ$, $\boldsymbol\kappa_s = \frac{2}{3d_s}\sum_{i\in s}(\bar u_s - u_i - o_i)(\cos\phi_i,\sin\phi_i)$, with 9 identifiable parameters: the pitch radius $d_s$ of each section and a zero offset $o_i$ per tendon, up to the per-section mean of the offsets, which cancels. They are identified by robust least squares on the 15 disc centres of all 4{,}400 training states. Gravity is not modelled. Tip and tilt come from the model frames, and the cloud is rendered by posing the robot model at the predicted frames in the same eight depth views.}

\subsection{Implementation of \method{}}
\label{sec:supp:toc_impl}
\label{sec:supp:impl}
\label{sec:supp:arch}
\rev{\Cref{tab:supp:hparams} lists the settings of \method{}.}

\begin{table*}[!tp]
  \centering
  \scriptsize %
  \setlength{\tabcolsep}{3pt}
  \caption{\rev{Hyperparameters and settings of \method{}.}}
  \label{tab:supp:hparams}
  \begin{minipage}[t]{0.495\textwidth}\vspace{0pt}
  \begin{tabular}{@{}>{\raggedright\arraybackslash}p{0.34\linewidth}>{\raggedright\arraybackslash}p{0.61\linewidth}@{}}
    \toprule
    Setting & Value \\
    \midrule
    \multicolumn{2}{l}{\emph{Observation model, Phase 1}} \\
    Model, initialization & \leavevmode ReSplat small, DL3DV checkpoint \\
    Input & \leavevmode $V{=}8$ context views, $448{\times}256$; $V{=}4$ on the real robot \\
    Gaussians & \leavevmode one per $4{\times}4$ patch, $7{,}168$ per view \\
    Refinement steps $J$ & \leavevmode $\{1,\dots,4\}$ random in training, 4 at inference \\
    Initialization / refinement & \leavevmode 20{,}000 / 3{,}000 steps \\
    Optimizer & \leavevmode AdamW, weight decay 0.01, one-cycle schedule, batch 1 \\
    Learning rate & \leavevmode $5\cdot10^{-5}$; depth backbone $2\cdot10^{-6}$ \\
    Photometric / perceptual & \leavevmode 1 / $\lambda_p=0.5$ \\
    Depth & \leavevmode $\lambda^{\mathrm{gm}}_{\mathrm{depth}}=10$, renderings in both stages and per-view depth predictions in the initialization \\
    Background opacity & \leavevmode $\lambda_{\mathrm{bg}}=1$, initialization only \\
    Anisotropy & \leavevmode $\lambda_a=10^{-3}$, $\kappa=10$, $\epsilon_s=0.1$\,mm, both stages \\
    Depth smoothness & \leavevmode $\lambda_{\mathrm{sm}}=0.01$, initialization only \\
    \midrule
    \multicolumn{2}{l}{\emph{Binding}} \\
    Thresholds & \leavevmode $\tau_o=0.3$, $d_b=2$\,mm \\
    Selection & \leavevmode \rev{grid $\tau_o\in\{0.3,0.5,0.7\}$, $d_b\in\{2,3,4\}$\,mm, lowest validation CD-full of the observation} \\
    \bottomrule
  \end{tabular}
  \end{minipage}\hfill
  \begin{minipage}[t]{0.495\textwidth}\vspace{0pt}
  \begin{tabular}{@{}>{\raggedright\arraybackslash}p{0.34\linewidth}>{\raggedright\arraybackslash}p{0.61\linewidth}@{}}
    \toprule
    Setting & Value \\
    \midrule
    \multicolumn{2}{l}{\emph{Back-projection and chain fit}} \\
    Opacity threshold & \leavevmode $\tau_\alpha=0.49$ by Otsu; 0.5 for the predicted cloud \\
    Voxel & \leavevmode 2\,mm \\
    Disc cylinders & \leavevmode $K{=}15$, radius 20\,mm, height 16\,mm; base cylinder $\mathcal{C}_0$ of radius 75.2\,mm; $K{=}9$ with the CAD dimensions on the real robot \\
    Points, penalty & \leavevmode $N\le4{,}000$; soft-$\ell_1$, $\delta=2$\,mm \\
    Smoothness, bound & \leavevmode $\lambda_s=0.005\sqrt{N}$; $\beta=0.6$\,rad \\
    Initializations & \leavevmode straight chain and the fit of the previous state; the lower mean distance is kept \\
    Staging & \leavevmode first 5, 10, 15 frames free, then all\fix{; 3, 6, 9 on the real robot} \\
    \midrule
    \multicolumn{2}{l}{\emph{Transition, Phase 2}} \\
    $f_\theta$ & \leavevmode MLP, SiLU, width 512, 3 hidden layers; inputs standardized by training statistics \\
    Optimizer, learning rate & \leavevmode Adam, $3\cdot10^{-4}$ \\
    Schedule & \leavevmode 40k iterations, batch 4, 500 warm-up iterations, $\times0.1$ at 80\,\% \\
    Training horizon & \leavevmode one step \\
    Loss weights & \leavevmode RGB 1, $\lambda^{\mathrm{tr}}_{\mathrm{depth}}=3$ \\
    Checkpoint & \leavevmode lowest validation Chamfer distance on the full cloud, the ACFM rule; the rule of \cref{sec:supp:protocol} on the real robot \\
    Seed & \leavevmode 0 in both phases \\
    \bottomrule
  \end{tabular}
  \end{minipage}
\end{table*}

\paragraph{\rev{Phase-1 objective.}}
\rev{Phase~1 keeps the two training stages of the feed-forward model~\cite{xu2026resplat}. The learning signal is the objective $\mathcal{L}_{\mathrm{gm}}$ of \cref{eq:init_loss}. The implementation applies its depth supervision under one weight to the rendered depth and to the per-view depth predictions of $\Phi_\psi$ and adds two standard regularizers. The regularizers carry no supervision, and their weights were held at one value in every Phase-1 run of this paper. The initialization stage trains the first prediction of $\Phi_\psi$ with
\mtgk{\begin{equation}
\begin{split}
  \mathcal{L}_{\mathrm{init}}&=\underbrace{\ell_1+\lambda_p\mathcal{L}_{\mathrm{perc}}+\lambda^{\mathrm{gm}}_{\mathrm{depth}}\mathcal{L}_{\mathrm{depth}}+\lambda_{\mathrm{bg}}\mathcal{L}_{\mathrm{bg}}}_{\mathcal{L}_{\mathrm{gm}}\text{ of \cref{eq:init_loss}}}+\mathcal{L}_{\mathrm{aux}},\\
  \mathcal{L}_{\mathrm{aux}}&=\lambda^{\mathrm{gm}}_{\mathrm{depth}}\mathcal{L}_{\mathrm{depth}}^{\mathrm{view}}+\underbrace{\lambda_a\mathcal{L}_{\mathrm{aniso}}+\lambda_{\mathrm{sm}}\mathcal{L}_{\mathrm{sm}}}_{\text{regularizers}},
\end{split}
\end{equation}}
where $\mathcal{L}_{\mathrm{depth}}$ and $\mathcal{L}_{\mathrm{bg}}$ are the terms of \cref{eq:depth_bg}, and $\mathcal{L}_{\mathrm{depth}}^{\mathrm{view}}$ applies the masked $\ell_1$ of $\mathcal{L}_{\mathrm{depth}}$ to the per-view depth predictions. The scale regularizer of the main text is the anisotropy term of PhysGaussian~\cite{xie2024physgaussian}, $\mathcal{L}_{\mathrm{aniso}}=\mathrm{mean}_i\max\!\big(0,\,s_{\max,i}/\max(s_{\min,i},\epsilon_s)-\kappa\big)$. It is zero for every Gaussian whose ratio of the largest to the smallest scale is at most $\kappa$, so it limits needle-shaped Gaussians and does not force isotropic ones. The smoothness regularizer $\mathcal{L}_{\mathrm{sm}}$ is the edge-aware smoothness of the per-view depth predictions in the released configuration of the feed-forward model, kept at its released weight. Neither regularizer uses a ground-truth depth map. In simulation the rendering was compared with the full training image without compositing, which is $\mathbf{c}=\mathbf{0}$ in \cref{eq:init_loss}. The refinement stage trains the recurrent update on its $J$ outputs with
\begin{equation}
  \mathcal{L}_{\mathrm{ref}}=\sum_{j=1}^{J}\gamma^{J-j}\Big(\ell_1^{(j)}+\lambda_p\mathcal{L}_{\mathrm{perc}}^{(j)}+\lambda^{\mathrm{gm}}_{\mathrm{depth}}\mathcal{L}_{\mathrm{depth}}^{(j)}+\lambda_a\mathcal{L}_{\mathrm{aniso}}^{(j)}\Big),
\end{equation}
with $\gamma=0.9$ and the first prediction frozen. The background term, the per-view depth supervision and the depth smoothness act in the initialization stage only. The $\ell_1$ and perceptual terms, the weights $\gamma^{J-j}$ and the depth smoothness are the released configuration of the feed-forward model, and the depth supervision, the background term and the anisotropy regularizer were added for \method{}.}

\paragraph{\mtgk{Loss ablation of Phase~1.}}
The four rows of \cref{tab:loss1}(a) were trained with one script and code version from the same pretrained weights, training images, seed and schedule of 20{,}000 initialization and 3{,}000 refinement steps, and the checkpoint rule selected refinement step 3{,}000 in every row. The rows differed only in $\lambda^{\mathrm{gm}}_{\mathrm{depth}}$ and $\lambda_{\mathrm{bg}}$, a removed term had its weight set to zero, and no other weight was retuned. A zero $\lambda^{\mathrm{gm}}_{\mathrm{depth}}$ removed the rendered-depth term of both stages together with $\mathcal{L}^{\mathrm{view}}_{\mathrm{depth}}$. Both regularizers were kept in all four rows, so row (i) was the rendering loss with these two regularizers and the table does not measure the separate effect of either.

\paragraph{Robot-specific inputs.}
\rev{The frame spacing, the body radius and a cylinder around the base are the robot-specific inputs, taken from the simulator model in simulation and from the CAD model on the real robot\mtgk{. The real robot adds the size of the marker cube from the CAD model and its calibrated mount offset and yaw of \cref{sec:supp:real}}. $\tau_\alpha$ is the Otsu threshold~\cite{otsu1979threshold} of the opacity histogram over the training views.}

\paragraph{Chain fit solver.}
\rev{\Cref{eq:icp_update} is minimized by a batched projected Levenberg--Marquardt solver. The penalty is $\rho(e)=\delta^{2}\big(\sqrt{1+(e/\delta)^{2}}-1\big)$, applied to the point distances and to the smoothness residuals $0.005\sqrt{N}\,(\bphi_{k}-\bphi_{k-1})$, where $N\le4{,}000$ is the number of fitted points, so the coefficient of \cref{eq:icp_update} is $\lambda_s=0.005\sqrt{N}$. Lengths are in metres, and $\delta=0.002$ applies to both residuals. In simulation, the four stages stop after at most 60, 60, 60 and 200 function evaluations, with cost and step tolerances of $10^{-6}$ in the first three stages and $10^{-8}$ in the last, and a gradient tolerance of $10^{-8}$ in all four. After the last stage, every frame beyond the last frame with at least 20 points within 3\,mm of its cylinder takes the bending of that frame. A state is fitted from the straight chain and from the fit of the previous state of its trajectory, and the fit with the lower mean distance is kept; the first state uses the straight chain alone. On the real test split the mean distances of the two starts differed by 0.0006\,mm on average and by at most 0.034\,mm. The transition network outputs the increment in radians from the difference of the commanded tendon lengths, and the commands are clipped to the command range in simulation. In simulation every cloud of every method and every reference of \cref{tab:supp:floor} was read with this solver and the same initialization rule.}

\paragraph{\mtgk{Frame reduction of \cref{fig:chain_k}.}}
\mtgk{\Cref{fig:chain_k} kept every $s$-th frame counted from the tip, and $s\in\{1,2,3,5\}$ gave 15, 8, 5 and 3 frames. The kept frames moved by the prediction of the reported model, and no network was retrained. Under rigid binding each Gaussian followed the kept frame with the nearest origin. This rule with all 15 frames moved the Gaussians by 0.014\,mm on average from their positions under the binding of \cref{eq:local}, and its CD and PSNR matched those of \cref{tab:main} at the reported precision. Under LBS each Gaussian blended the transforms of the two kept frames that enclose its disc with the weights $1-t$ and $t$, where $t\in[0,1]$ is the projection of its centre on the segment between the origins of these two frames. A Gaussian below the lowest kept frame followed that frame under both rules, and the Gaussians of the base stayed fixed.}

\paragraph{Binding at the base.}
\mtgk{A Gaussian was kept when its opacity was at least $\tau_o$ and its signed distance to the cylinder it was bound to was at most $d_b$, so a Gaussian inside a cylinder was kept.} \fix{On the real robot, a Gaussian within 2\,mm of the box of the marker cube that is nearer to the box than to every ring cylinder and to the base is bound to the last frame.}

\paragraph{Transition designs of \cref{tab:transition}.}
The three rows were trained with the data, schedule, seed and checkpoint rule of \method{} and differ in the output of the network alone. The free SE(3) row outputs a separate rigid motion for each frame from the bending state and the command change, so the frames are not held on a chain. The linear row outputs a Jacobian from the bending state and multiplies it by the command change. Both rows used the width, depth weight and schedule of \method{}.

\paragraph{Increments of \cref{tab:increment}.}
The chain fit reads the cloud that PCC or ACFM predicts at the previous and at the next command, and the difference of the two fitted bending states is the increment. This increment is added to the backbone fitted to the observation of \method{} and moves the same Gaussians. \Cref{tab:supp:real_extra} gives the same rows with all metrics on both datasets.

\subsection{Computational Cost}
\label{sec:supp:cost}
\paragraph{Training and memory.}
In simulation the initialization stage of Phase~1 took 1.7\,h, its refinement stage 0.5\,h and Phase~2 1.3\,h on one B200 GPU. \method{} was trained once for each robot, and no network weight was updated at test time. The peak GPU memory reported by nvidia-smi was 15.7\,GiB in the initialization stage and 12.8\,GiB in the refinement stage of Phase~1, 1.1\,GiB in Phase~2 and 5.7\,GiB for the reconstruction at inference.

\paragraph{Complexity.}
With $V$ views the reconstruction is one forward pass of the feed-forward model and returns $G=7{,}168\,V$ Gaussians. One iteration of the chain fit evaluates the distance of each of the $N\le4{,}000$ points to each of the $K$ cylinders and forms the normal equations of the $2K$ bending components, at a cost of $O(NK^{2}+K^{3})$. The binding assigns each Gaussian to one of the $K$ cylinders in $O(GK)$. The transition is one pass of an MLP whose cost does not depend on $G$, and the motion applies one rigid transform to each Gaussian in $O(G)$. A further command on the same observation therefore costs $O(G)$ and the rendering. Of the 0.83\,s of one prediction in \cref{fig:time}, the reconstruction took 0.22\,s, the chain fit 0.39\,s, the binding 0.12\,s and the transition with the motion of the Gaussians 0.003\,s; the remaining 0.10\,s went to the other steps of the pipeline.

\paragraph{Timing protocol.}
\label{sec:supp:timing}
\rev{All times of \cref{fig:time} are medians over the 50 test transitions of trajectories 0 to 4, steps 1 to 10, measured on one NVIDIA B200 GPU with batch size one, five warm-up transitions and GPU synchronization around every stage. The tip of the point-cloud baselines was read by the GPU backbone fit of \cref{sec:supp:icp} under the same initialization rule, which adds 0.30 to 0.65\,s to their totals. PCC rendered its depth views on the CPU. The MuJoCo reference is the settling time of one transition under the data-generation rule of \cref{sec:supp:data} with one thread on the CPU of the same machine (median 4.11\,s, mean 7.36\,s).}

\section{Extended Results of Next-State Prediction}
\label{sec:supp:additional}
\subsection{Qualitative Results}
\label{sec:supp:toc_qual}
\label{sec:supp:qual}
\mtgk{Figures~\ref{fig:supp:qual_sim_best} and~\ref{fig:supp:qual_sim_worst} show test transitions of ACFM-Transition and Figures~\ref{fig:supp:qual_real_best} and~\ref{fig:supp:qual_real_worst} of the real robot, for all methods. The transitions were taken at every fifth percentile of the tip error of \method{}, from the 5th to the 25th for the lowest errors and from the 80th to the 100th for the highest, over the 500 test transitions of ACFM-Transition and the 996 of the real robot. The 100th percentile is the transition with the largest tip error.}

\subsection{Reference Rows for the Analysis of the Learned Transition}
\label{sec:supp:refrows}
\Cref{tab:supp:real_extra} adds two kinds of rows to \cref{tab:main}, and each answers one question about the learned transition.

\par\medskip\noindent\begin{minipage}{\columnwidth}
  \centering
  \scriptsize %
  \setlength{\tabcolsep}{1.15pt}
  \setlength{\abovecaptionskip}{2pt}
  \captionof{table}{\mtgk{Reference rows for \method{} on the test transitions.}}
  \label{tab:supp:real_extra}
  \begin{tabular}{@{}l cc ccc cc@{}}
    \toprule
    & CD$\downarrow$ & EMD$\downarrow$ & PSNR$\uparrow$ & SSIM$\uparrow$ & LPIPS$\downarrow$ & Pos.$\downarrow$ & Tilt$\downarrow$ \\
    \midrule
    \multicolumn{8}{@{}l}{\emph{ACFM-Transition, 500 transitions}} \\
    Zero transition & 1.52 & 0.737 & 15.78 & 0.762 & 0.074 & 22.6 & 9.07 \\
    \method{} without transition & 1.59 & 0.832 & 15.67 & 0.745 & 0.077 & 22.7 & 9.19 \\
    PCC increment & 0.149 & 0.354 & 23.42 & 0.922 & 0.017 & 3.23 & 1.21 \\
    ACFM increment & 0.102 & 0.333 & 23.34 & 0.927 & 0.016 & 3.59 & 5.20 \\
    \method{} & 0.085 & 0.323 & 24.74 & 0.941 & 0.013 & 1.47 & 1.15 \\
    \midrule
    \multicolumn{8}{@{}l}{\emph{Real robot, 996 transitions}} \\
    Zero transition & 0.117 & 0.260 & 14.18 & 0.790 & 0.136 & 4.52 & 2.20 \\
    \method{} without transition & 0.193 & 1.190 & 15.21 & 0.805 & 0.106 & 5.34 & 4.52 \\
    PCC increment & 0.177 & 1.175 & 15.44 & 0.810 & 0.105 & 4.76 & 4.41 \\
    ACFM increment & 0.496 & 1.055 & 13.91 & 0.783 & 0.122 & 13.6 & 8.69 \\
    \method{} & 0.159 & 1.176 & 15.83 & 0.817 & 0.101 & 3.76 & 4.16 \\
    \bottomrule
  \end{tabular}
\end{minipage}\par\medskip

\paragraph{Effect of the transition.} \method{} without transition keeps the reconstruction of the observed state as its prediction. In simulation, the learned transition improved all seven metrics over this row and lowered the tip error from 22.7 to 1.47\,mm. On the real robot, it also improved all seven metrics and lowered the tip error from 5.34 to 3.76\,mm.

\paragraph{Increments of the baselines.} The PCC and ACFM increment rows extend \cref{tab:increment} to all metrics and to the real robot: the increment of each baseline moves the same observed Gaussians and backbone. In simulation, \method{} was the best of the three in all seven metrics, with a tip error of 1.47\,mm against 3.23\,mm and 3.59\,mm. On the real robot, \method{} had a lower CD, tip error and tilt than both rows, with a tip error of 3.76\,mm against 4.76\,mm and 13.6\,mm, and the paired bootstrap intervals of these differences exclude zero. Its EMD did not differ from that of the PCC increment, and the ACFM increment had the lowest EMD of the three.

\subsection{Transitions over Several Recorded Steps on the Real Robot}
\label{sec:supp:kstep}
\mtgk{\Cref{fig:motion_zero}(b) applied the one-step model $k$ times. Here the transition network was trained for each $k$ on real pairs of states $k$ recorded steps apart and predicted the second state of a pair in one application. \Cref{fig:supp:kstep} shows that the learned transition had the lowest tip error at every $k$ and the lowest CD at $k=2$, 3 and 5, and that its gain in tip error over the zero transition grew with $k$.}

\paragraph{Setting.} \mtgk{A pair holds two settled states of one trajectory. \method{} received the observation of the first state and the command change between the two states, and no method received a command of the states between them. Cho~\etal\ was trained on the same pairs, and both used their reported settings. The PCC and ACFM increments of \cref{tab:supp:real_extra} take the absolute commands of the two states and were applied without retraining. The points at $k=1$ are the reported models.}

\paragraph{Result.} \mtgk{The zero transition keeps the observed state as the prediction. The tip error of \method{} was lower than that of the zero transition by 0.8\,mm at $k=1$ and by 3.1, 5.4 and 8.7\,mm at $k=2$, 3 and 5, and the intervals of these differences exclude zero. The zero transition had the lower CD at $k=1$ as in \cref{tab:main}, and \method{} had the lower CD at $k=2$, 3 and 5. Among all rows at $k=2$, 3 and 5, \method{} had the lowest CD and tip error and the best appearance in all three metrics, and the intervals of its differences in CD and tip error from every other row exclude zero. The intervals of the differences in CD and tip error between the trained model and the one-step model applied $k$ times include zero at $k=2$, 3 and 5. As in \cref{tab:main,tab:supp:real_extra}, the zero transition, the ACFM increment and Cho~\etal\ had a lower EMD, and the tilt of the zero transition was lower at $k=2$ and higher at $k=3$ and $k=5$.}

\par\medskip\noindent\begin{minipage}{\columnwidth}
  \centering
  \includegraphics[width=\columnwidth]{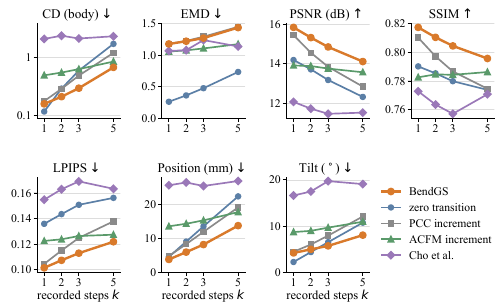}
  \setlength{\abovecaptionskip}{2pt}
  \captionof{figure}{\mtgk{Real-robot test pairs of settled states $k$ recorded steps apart.}}
  \label{fig:supp:kstep}
\end{minipage}\par\medskip

\subsection{Training Seeds}
\label{sec:supp:stat}
\fix{The transition network was trained with two more seeds in simulation, with Phase~1 and the checkpoint rule unchanged. The three runs differed by at most 0.0007 in CD, 0.04\,dB in PSNR and 0.02\,mm in tip position, and the margin to the best baseline was 0.030 in CD.}

\begin{figure*}[!tb]
  \centering
  \includegraphics[width=0.96\textwidth]{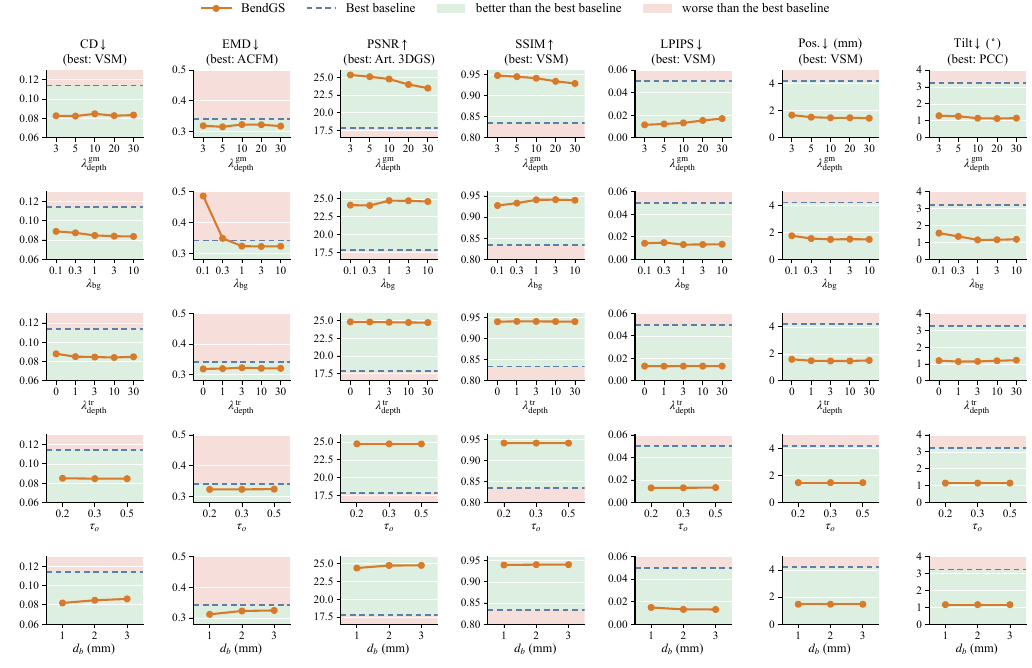}
  \caption{\rev{All metrics of \cref{tab:main} on the ACFM-Transition test set with one hyperparameter changed at a time. Rows from the top: depth weight of Phase~1, $\lambda_{\mathrm{bg}}$, depth weight of Phase~2, $\tau_o$ and $d_b$. Dashed: best baseline; green and red: the sides better and worse than it.}}
  \label{fig:supp:sens_test}
\end{figure*}
\subsection{Tip Pose with the Shared Readout}
\label{sec:supp:toc_realres}
\label{sec:supp:real_results}

\fix{\method{} had the lowest tip error and tilt among the compared methods on the real robot under either readout of its tip pose. \Cref{tab:main} reports the tip pose of \method{} from its predicted backbone, without any fit. On the real robot the tip point lies on the axis of the last frame at the mount offset $\ell_m=23$\,mm of the cube, which was calibrated without labels,
\begin{equation}
\begin{aligned}
  \mathbf{p}^{\mathrm{tip}}_{t+1}&=\mathbf{t}_K(\bphi_{t+1})+\ell_m\,\bR_K(\bphi_{t+1})\,\mathbf{e}_z,\\
  \mathbf{n}^{\mathrm{tip}}_{t+1}&=\bR_K(\bphi_{t+1})\,\mathbf{e}_z,
\end{aligned}
  \label{eq:supp:tip_real}
\end{equation}
which is \cref{eq:tip} with the mount counted as one more straight link after frame $K$. No image or cloud of the next state enters this output. \Cref{tab:supp:native} applies the shared chain fit of the baselines to every method, and \method{} stayed first in both columns, with 6.09\,mm against the 12.1\,mm of NJF and $4.89^\circ$ against the $16.3^\circ$ of ACFM. In simulation the two readouts of \method{} agreed within 0.02\,mm and $0.03^\circ$, as \cref{sec:supp:icp} reports.}
\par\medskip\noindent\begin{minipage}{\columnwidth}
  \centering
  \scriptsize %
  \setlength{\tabcolsep}{4pt}
  \setlength{\abovecaptionskip}{2pt}
  \captionof{table}{\fix{Tip pose on the real robot (test) with the shared chain fit applied to every method. Bold: best; underline: second. $^\dagger$As in \cref{tab:main}.}}
  \label{tab:supp:native}
  \begin{tabular}{@{}l cc cc@{}}
    \toprule
    & \multicolumn{2}{c}{\cref{tab:main}} & \multicolumn{2}{c}{Shared chain fit} \\
    \cmidrule(lr){2-3}\cmidrule(lr){4-5}
    Method & Pos.$\downarrow$ & Tilt$\downarrow$ & Pos.$\downarrow$ & Tilt$\downarrow$ \\
    \midrule
    PCC & 45.9 & 35.8 & 45.9 & 32.6 \\
    VSM & 35.8 & 23.1 & 35.8 & 23.1 \\
    Cho~\etal & 25.7 & 16.6 & 25.7 & 16.6 \\
    FFKSM & 133.7 & 148.6 & 133.7 & 148.6 \\
    Articulated 3DGS & 33.6 & 17.1 & 33.6 & 17.1 \\
    ACFM & 24.4 & \underline{16.3} & 24.4 & \underline{16.3} \\
    NJF$^\dagger$ & \underline{12.1} & 21.8 & \underline{12.1} & 21.8 \\
    \rowcolor{oursbg}\method{} & \textbf{3.76} & \textbf{4.16} & \textbf{6.09} & \textbf{4.89} \\
    \bottomrule
  \end{tabular}
\end{minipage}\par\medskip

\subsection{Empty Predictions of FFKSM}
\label{sec:supp:ffksm_empty}
\rev{\method{} kept its lead on the transitions where FFKSM gave a non-empty prediction. FFKSM returned no point above its density threshold in the body region for 267 of the 996 real test transitions, and its prediction was non-empty on the others. \Cref{tab:main} averages FFKSM over these non-empty transitions, 729 for the CD and EMD and 693 for the tip pose, and every other method over all 996. The shared readout requires 50 points of a cloud inside its fitting region, and 36 of the 729 non-empty predictions held fewer. The appearance of FFKSM is averaged over all 996 transitions, since an empty prediction still renders the background. \Cref{tab:supp:ffksm_subset} scores every method on the same non-empty transitions only. The first place in each column did not change: \method{} stayed first in CD, tip position and tilt, and ACFM in EMD. The density threshold was selected on the simulation validation split and kept on the real robot.}

\par\medskip\noindent\begin{minipage}{\columnwidth}
  \centering
  \scriptsize %
  \setlength{\tabcolsep}{3pt}
  \setlength{\abovecaptionskip}{2pt}
  \captionof{table}{\rev{Real-robot test transitions with a non-empty FFKSM prediction (729 for CD and EMD, 693 for tip pose). Tip pose of \method{} and PCC from their own frames. Bold: best; underline: second. $^\dagger$As in \cref{tab:main}.}}
  \label{tab:supp:ffksm_subset}
  \begin{tabular}{@{}l cccc@{}}
    \toprule
    Method & CD$\downarrow$ & EMD$\downarrow$ & Pos.$\downarrow$ & Tilt$\downarrow$ \\
    \midrule
    PCC & 5.16 & 1.77 & \fix{34.1} & \fix{32.5} \\
    VSM & 3.39 & 2.95 & 33.1 & 20.3 \\
    Cho~\etal & 2.02 & \underline{1.01} & 25.1 & \underline{14.5} \\
    FFKSM & 39.1 & 6.26 & 133.7 & 148.6 \\
    Articulated 3DGS & 2.76 & 3.59 & 31.0 & 14.7 \\
    ACFM & \underline{1.95} & \textbf{0.938} & 23.7 & 15.3 \\
    NJF$^\dagger$ & 5.65 & 1.83 & \underline{11.5} & 20.9 \\
    \rowcolor{oursbg}\method{} & \textbf{0.153} & 1.155 & \textbf{\fix{3.89}} & \textbf{\fix{3.76}} \\
    \bottomrule
  \end{tabular}
\end{minipage}\par\medskip

\section{Sensitivity to Hyperparameters}
\label{sec:supp:toc_sens}
\label{sec:supp:sens}
\fix{\Cref{fig:supp:sens_test} extends \cref{fig:sens} to all seven metrics of \cref{tab:main} and adds the two binding thresholds. All five sweeps were run on the test set after every setting and checkpoint of the reported model was frozen, and these test values took no part in any choice. The body CD varied by at most 0.006 within each sweep of the three loss coefficients and by at most 0.005 within each sweep of the two binding thresholds. All seven metrics stayed better than the best baseline of each metric in every setting, with one exception: the EMD at $\lambda_{\mathrm{bg}}$ of 0.1 and 0.3 was 0.486 and 0.348 against 0.341 for ACFM.}\looseness=-1

\clearpage
\begin{figure*}[p]
  \centering
  \includegraphics[width=0.78\textwidth]{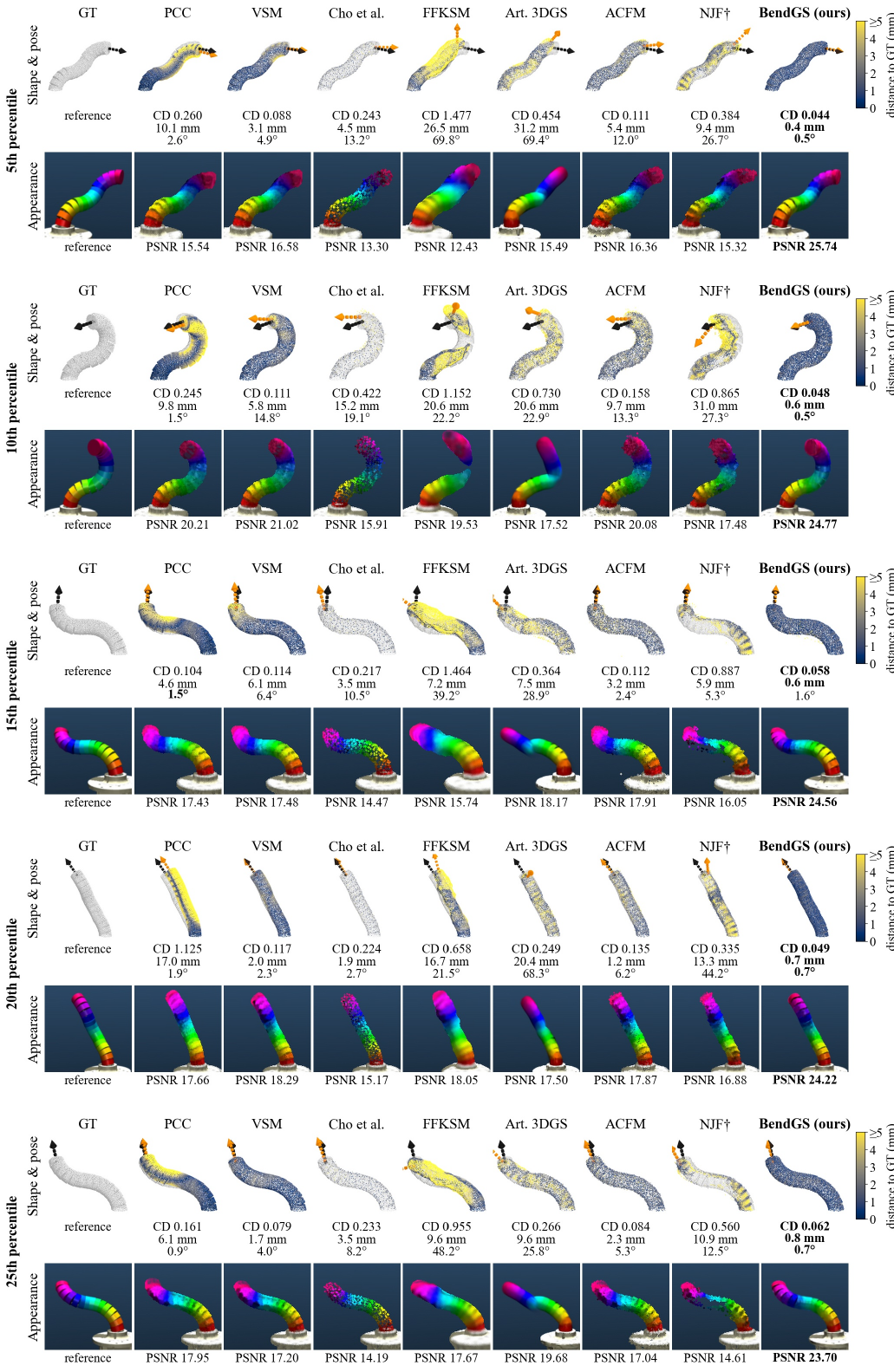}
  \caption{\mtgk{Test transitions of ACFM-Transition with low tip errors of \method{}, the 5th to 25th percentiles, in the layout of \cref{fig:main_qual}. Bold: best.}}
  \label{fig:supp:qual_sim_best}
\end{figure*}
\begin{figure*}[p]
  \centering
  \includegraphics[width=0.78\textwidth]{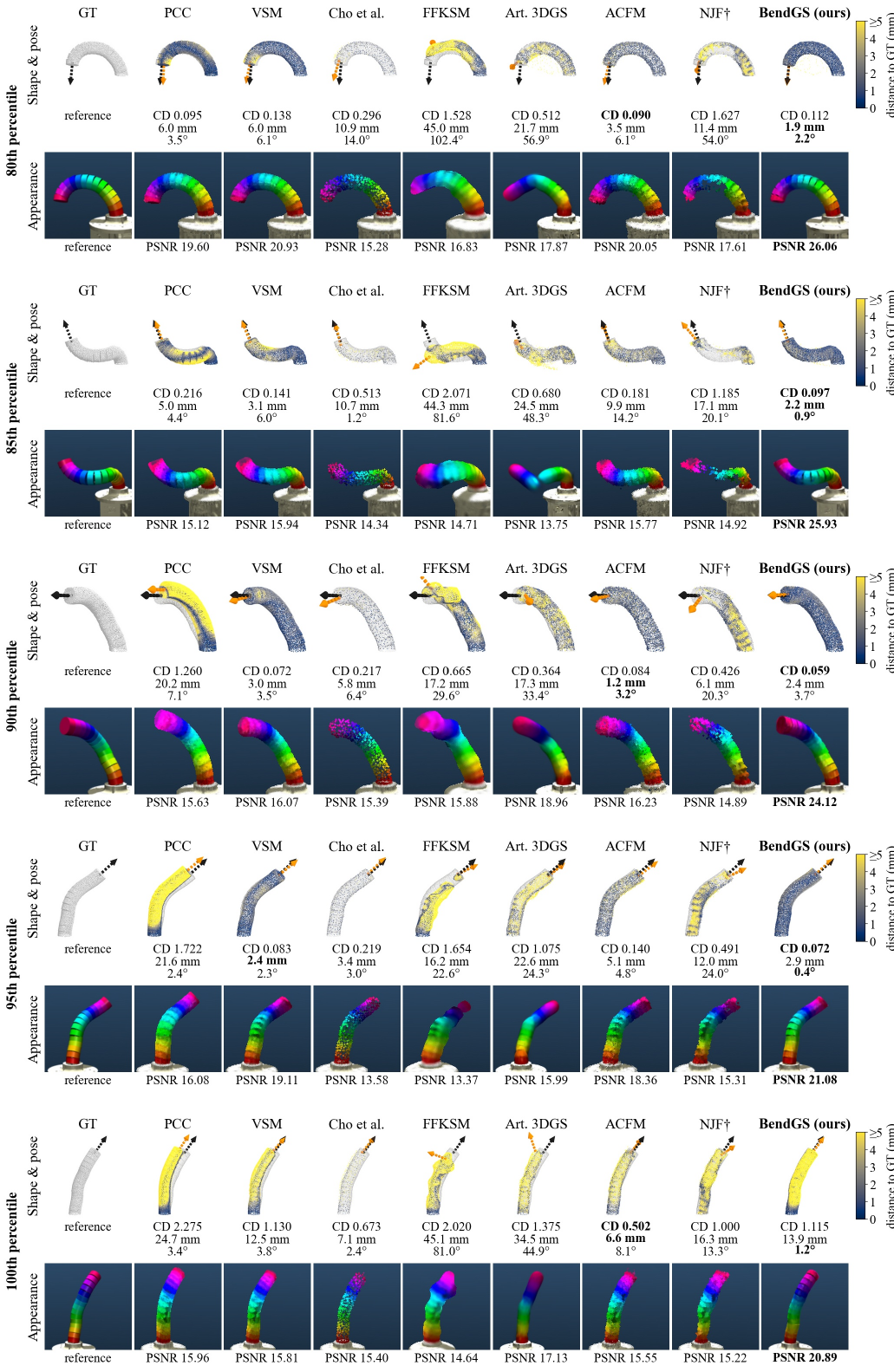}
  \caption{\mtgk{Test transitions of ACFM-Transition with the highest tip errors of \method{}, the 80th to 100th percentiles, in the layout of \cref{fig:main_qual}. Bold: best.}}
  \label{fig:supp:qual_sim_worst}
\end{figure*}
\begin{figure*}[p]
  \centering
  \includegraphics[width=0.78\textwidth]{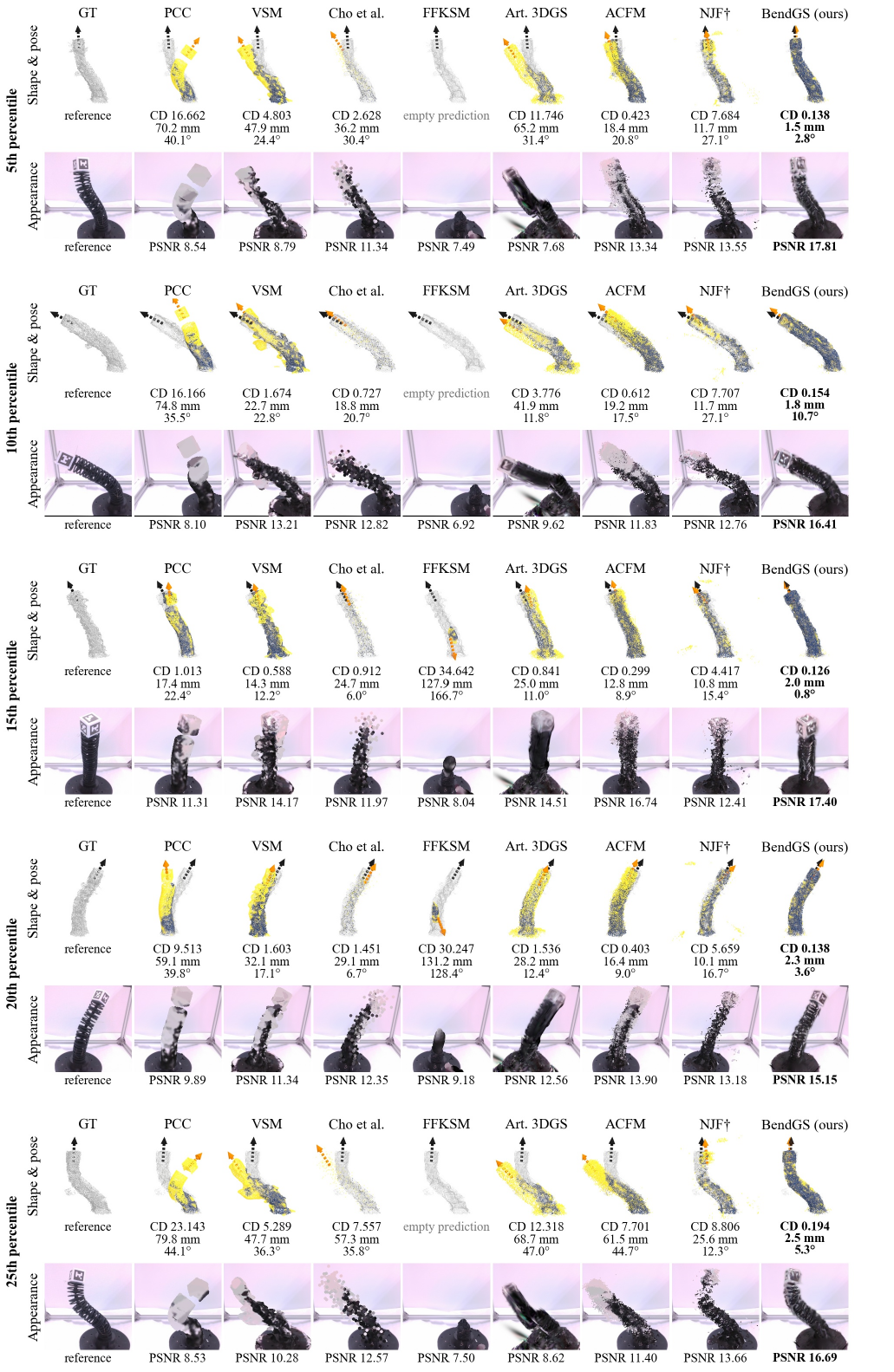}
  \caption{\mtgk{Test transitions of the real robot with low tip errors of \method{}, the 5th to 25th percentiles, in the layout of \cref{fig:main_qual}. Bold: best.}}
  \label{fig:supp:qual_real_best}
\end{figure*}
\begin{figure*}[p]
  \centering
  \includegraphics[width=0.78\textwidth]{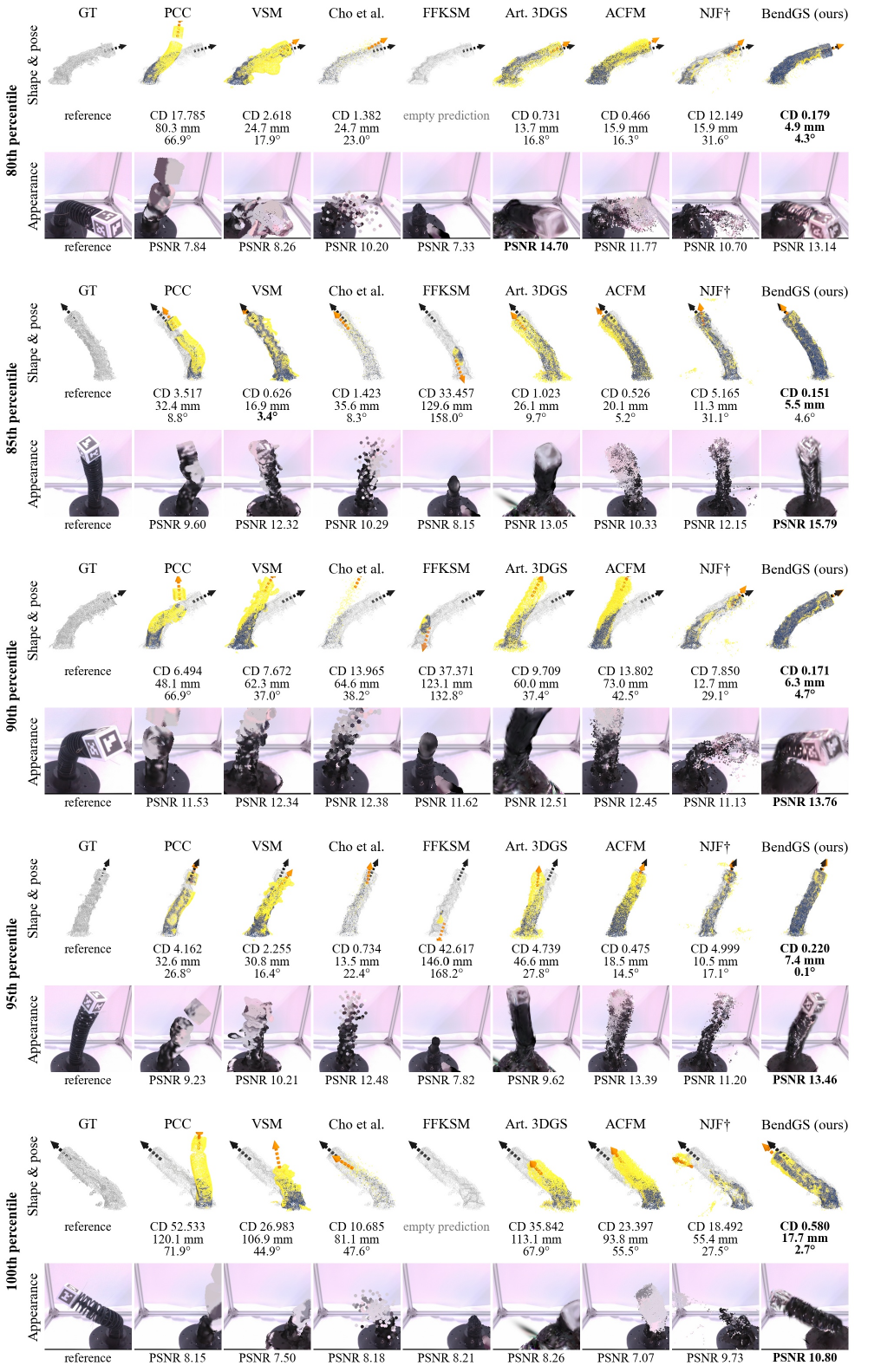}
  \caption{\mtgk{Test transitions of the real robot with the highest tip errors of \method{}, the 80th to 100th percentiles, in the layout of \cref{fig:main_qual}. Bold: best.}}
  \label{fig:supp:qual_real_worst}
\end{figure*}

\clearpage
{
    \small
    \bibliographystyle{ieeenat_fullname}
    \bibliography{main}
}

\end{document}